%% file: main.tex
\documentclass{article}

\usepackage[preprint]{corl_2026} 

\usepackage[utf8]{inputenc} 
\usepackage[T1]{fontenc}    
\usepackage{url}            
\usepackage{booktabs}       
\usepackage{amsfonts}       
\usepackage{nicefrac}       
\usepackage{microtype}      

\usepackage{amsthm}
\usepackage{multirow}
\usepackage{array}
\usepackage{placeins}
\usepackage{adjustbox}
\usepackage[table]{xcolor}
\definecolor{commentgray}{rgb}{0.5,0.5,0.5}
\definecolor{lightgray}{gray}{0.8}
\usepackage{amsmath,amssymb}
\usepackage[skins,breakable]{tcolorbox}
\usepackage{empheq}
\usepackage{xspace}
\usepackage{float}
\usepackage{enumitem}
\usepackage{caption}
\usepackage{subcaption}
\usepackage{wrapfig}
\usepackage{tabularx}
\usepackage{algorithm}
\usepackage{algpseudocode}

\setlist[itemize]{topsep=-1pt, itemsep=1pt, parsep=0pt, partopsep=0pt,left=1pt}

\definecolor{rowgray}{gray}{0.96}

\newcommand\blfootnote[1]{%
  \begingroup
  \renewcommand\thefootnote{}\footnote{#1}%
  \addtocounter{footnote}{-1}%
  \endgroup
}

\title{PHASER: Phase-Aware and Semantic Experience Replay for Vision-Language-Action Models}

\author{
  \bfseries Ziyang Chen$^{1,\dagger}$, \enspace Shaoguang Wang$^{1,\dagger}$, \enspace Weiyu Guo$^{1,*}$, \enspace Qianyi Cai$^{1}$, \enspace He Zhang$^{1}$, \\[2pt]
  \bfseries Pengteng Li$^{1}$, \enspace Yiren Zhao$^{1}$, \enspace Yandong Guo$^{2}$ \\[6pt]
  \normalfont $^{1}$Thrust of AI, HKUST(Guangzhou), Guangzhou, China \\[2pt]
  $^{2}$AI\textsuperscript{2} Robotics, Shenzhen, China
}

\begin{document}
\maketitle

\blfootnote{$^{\dagger}$Equal contribution.}
\blfootnote{$^{*}$Corresponding author: \texttt{guoweyu96@gmail.com}}

\input{sec/abstract}


\input{sec/intro}
\input{sec/Method}
\input{sec/Experiment}

\input{sec/Analysis}

\input{sec/relatedwork}

\input{sec/limitations}
\input{sec/conclusion}

\bibliography{references}  

\clearpage
\appendix
\input{sec/appendix.tex}

\end{document}

%% file: sec/abstract.tex



\begin{abstract}
Vision-Language-Action (VLA) models have achieved remarkable success in language-conditioned robotic manipulation. However, deploying these models in open-ended environments requires continuously acquiring novel skills, a process that inevitably triggers severe catastrophic forgetting of previously learned behaviors. While experience replay (ER) serves as a standard mitigating strategy, naive uniform sampling fundamentally misaligns with the temporal characteristics of manipulation trajectories. It systematically under-samples brief but causally critical sub-skills, leading to \emph{phase starvation}, and completely overlooks the varying degrees of forgetting across historical tasks. To overcome these limitations, we introduce \textbf{Phaser}, an architecture-agnostic continual learning framework. Phaser employs a phase-centric capacity allocation to guarantee equal memory support for all sub-skills, coupled with a multi-modal interference routing strategy that dynamically prioritizes historical phases at high risk of forgetting. Furthermore, to enable fully autonomous lifelong adaptation, we integrate \emph{Auto-PC}, a lightweight pipeline combining unsupervised action-signal change-point detection with VLM-based semantic verification to extract temporal boundaries without intensive manual supervision. Evaluated across three VLA backbones on LIBERO continual learning suites, Phaser yields substantial empirical improvements, increasing Average Success Rate (ASR) by up to $31\%$ over matched-budget ER and achieving an $87.8\%$ final ASR on LIBERO-Goal CL setting.
\end{abstract}

%% file: sec/intro.tex
\section{Introduction}
\label{sec:intro}

\begin{figure*}[t]
    \centering
    \includegraphics[width=\textwidth]{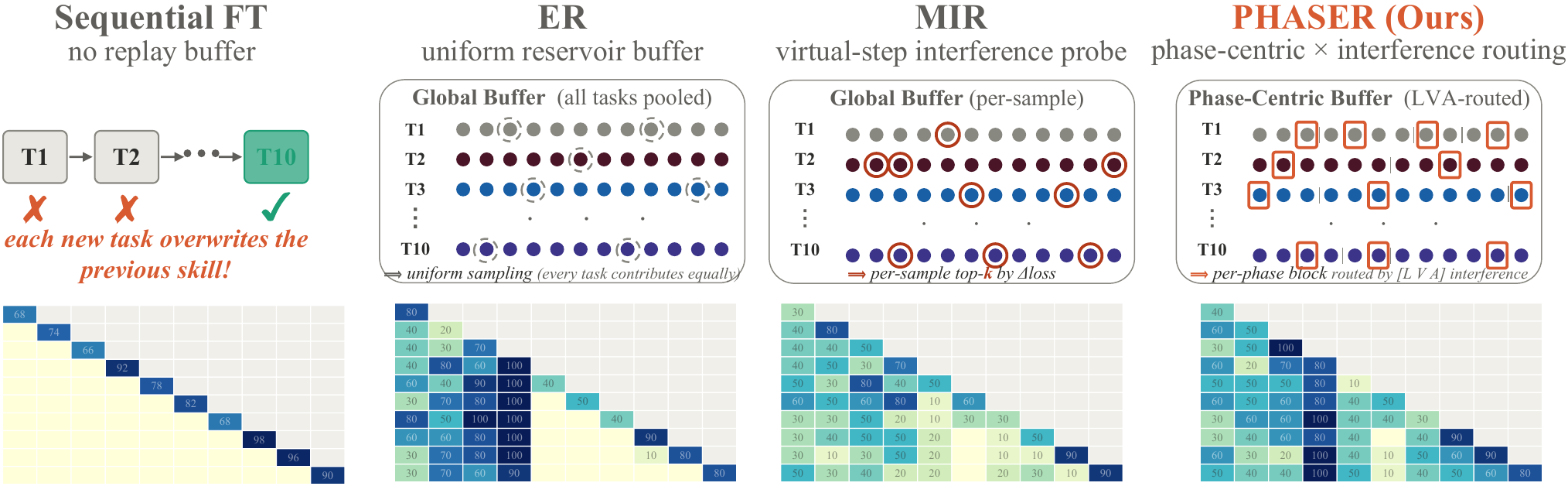}
    \caption{\textbf{What PHASER does and why it differs from standard experience replay.} A VLA agent learns a stream of language-conditioned manipulation tasks; each demonstration trajectory decomposes into temporally extended sub-skills (\emph{approach}, \emph{grasp}, \emph{transport}, $\dots$). The four columns contrast replay schemes under a matched budget. \emph{Sequential FT} keeps no replay buffer, so each new task overwrites the previously acquired skill. \emph{ER} samples uniformly from a global reservoir, allocating frame-level slots in proportion to phase duration---under-supporting brief but causally critical sub-skills (\emph{phase starvation}) and treating all historical tasks as interchangeable. \emph{MIR} retrieves per-sample top-$k$ exemplars by virtual-step loss increase, but still selects at the frame level over a global buffer. \emph{PHASER} (Ours, rightmost) replaces these choices with two structural priors: a \emph{phase-centric buffer} that guarantees equal frame support per atomic sub-skill, and a \emph{multi-modal interference router} that concentrates replay on historical phases sharing perceptual context with the current task but requiring divergent kinematics. The result is a single drop-in replay scheme that significantly improves retention over matched-budget ER across three VLA backbones and two LIBERO suites, without any architectural change to the policy.}
    \label{fig:teaser}
\end{figure*}

Vision-Language-Action (VLA) models~\citep{brohan2023rt2, kim2024openvla, black2024pi_0} have achieved remarkable success in language-conditioned robotic manipulation. Deploying these models in open-ended environments requires them to continuously acquire new skills, a process that inevitably triggers catastrophic forgetting of previously learned behaviors~\citep{mccloskey1989catastrophic}. To mitigate this vulnerability, experience replay (ER)~\citep{rolnick2019experience} serves as a simple yet effective method to preserve the foundational capabilities of pretrained VLAs~\citep{liu2026pretrained}.

However, standard ER relies on a uniform sampling strategy that fundamentally misaligns with the temporal characteristics of robotic manipulation. First, it ignores the temporal non-uniformity inherent in manipulation trajectories. Because a complete task naturally decomposes into distinct phases (e.g., \emph{approach}, \emph{grasp}, and \emph{transport}) that vary drastically in duration, sub-actions are distributed highly unevenly over time. Consequently, purely random or uniform frame sampling inherently allocates buffer capacity proportional to phase length, severely under-sampling brief but causally critical sub-skills. This phenomenon, which we define as \emph{phase starvation}, disproportionately degrades end-to-end task success. Second, uniform ER treats all historical tasks identically, allocating an equal replay budget to each past task. This strategy completely overlooks the fact that different historical tasks experience varying degrees of forgetting during the acquisition of a new skill. In manipulation, task interference is highly non-uniform: while some tasks share reusable sub-skills, others share overlapping perceptual contexts yet demand entirely different physical actions, posing a significantly higher risk of forgetting specific historical behaviors.

To overcome the limitations of naive uniform sampling, we introduce \textbf{PHASER} (\textbf{PH}ase-\textbf{A}ware and \textbf{S}emantic \textbf{E}xperience \textbf{R}eplay), a novel continual learning framework that operates on two complementary levels. To resolve intra-task phase starvation, PHASER enforces a \emph{Phase-Centric Capacity Allocation}. By shifting the fundamental unit of memory from the overall task down to the individual phase, it assigns an equal frame budget to every sub-skill regardless of its temporal duration, directly ensuring that brief but critical actions are adequately learned. Furthermore, to prevent inter-task budget misallocation and non-uniform forgetting, the framework employs \emph{Multi-Modal Interference Routing}. By scoring historical phases against the current task using a tri-modal (Language, Vision, Action) prototype distance, it intelligently routes the replay budget toward past phases that share high perceptual context but require divergent physical actions. Crucially, because this routing is evaluated only once per task transition, it adds only marginal computational overhead. Finally, because phase-centric memory relies on temporal boundaries, we enable fully autonomous lifelong learning by integrating \textbf{Auto-PC} (Phase Candidates). Since directly prompting Vision-Language Models (VLMs) to segment videos empirically yields temporally imprecise boundaries~\citep{zhang2023gpt}, Auto-PC couples unsupervised action-signal change-point detection with a single VLM semantic verification step. This lightweight pipeline extracts and labels sub-skills automatically, securing robust downstream performance without relying on external supervision.


In summary, our primary contributions are:
\begin{enumerate}[label=(\arabic*), leftmargin=1.5em, nosep]
    \item We propose \textbf{PHASER}, a novel, architecture-agnostic continual learning framework for Vision-Language-Action (VLA) models. By explicitly modeling the temporal structure of manipulation trajectories, PHASER systematically overcomes the inherent limitations of standard uniform experience replay: intra-task phase starvation and inter-task budget misallocation.
    \item We employ two core mechanisms within PHASER: a phase-centric capacity allocation that guarantees equal memory support for all sub-skills, and a multi-modal interference routing strategy that dynamically prioritizes historical phases at high risk of forgetting. Additionally, to enable fully autonomous operation, we introduce \textbf{Auto-PC}, an automated pipeline coupling unsupervised action-signal change-point detection with VLM-based semantic verification.
    \item We extensively evaluate PHASER across three VLA backbones (OpenVLA-7B, QwenGR00T-3B, QwenOFT-3B) and two LIBERO suites. Under strict memory constraints, our framework yields substantial empirical improvements, increasing the Average Success Rate (ASR) by up to $31\%$ over matched-budget ER and achieving an $87.8\%$ final ASR on the LIBERO-Goal CL setting.
\end{enumerate}

%% file: sec/Method.tex
\section{Method}
\label{sec:method}

\begin{figure*}[t]
    \centering
    \includegraphics[width=\textwidth]{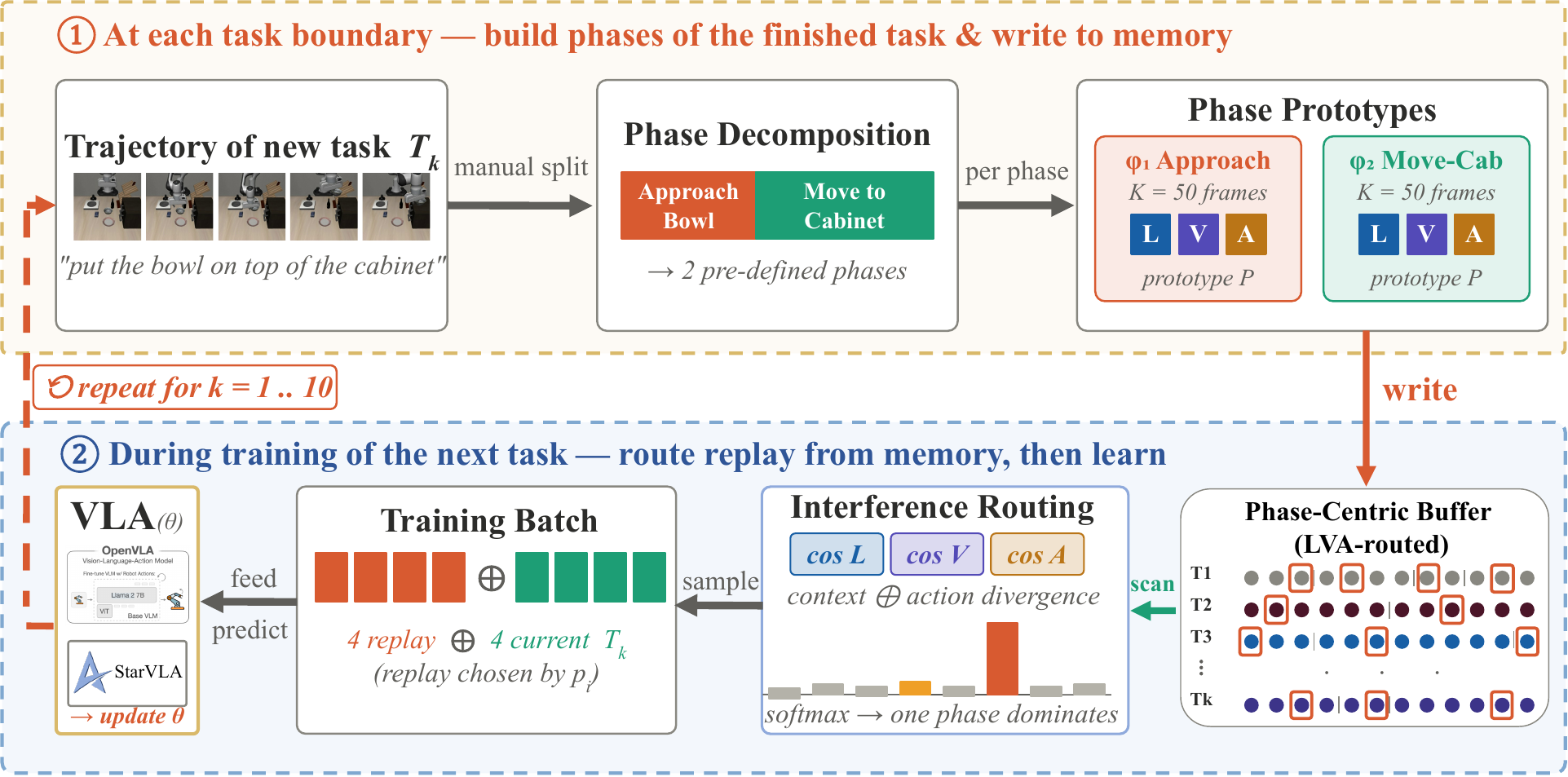}
    \caption{\textbf{PHASER pipeline overview.} At each task boundary, trajectories are partitioned by phase annotation; each phase writes a fixed-size bucket (\emph{intra-task} allocation, \S\ref{sec:intra_task}). When training the next task, a tri-modal $(L,V,A)$ prototype score $U_{i,k}$ ranks historical phases by interference risk, and a Boltzmann distribution $p_{i,k}{=}\mathrm{softmax}(U_{i,k}/\tau)$ drives replay sampling (\emph{inter-task} routing, \S\ref{sec:inter_task}). Routing is computed once per transition, so per-step replay cost matches vanilla ER.}
    \label{fig:pipeline}
\end{figure*}

We reinterpret VLA continual learning through a Semi-Markov Decision Process (SMDP) lens (Sec.~\ref{sec:formulation}) to expose the structural mismatch between trajectory data and uniform frame-level replay. PHASER addresses it via two data-side principles: (i) intra-task \textit{Phase-Centric Capacity Allocation} that equalises buffer support across sub-skills (Sec.~\ref{sec:intra_task}), and (ii) inter-task \textit{Multi-Modal Interference Routing} that concentrates rehearsal on historical phases most at risk of being overwritten (Sec.~\ref{sec:inter_task}). Phase boundaries come from human annotation (paper default) or an automatic VLM-verified pipeline (Sec.~\ref{sec:autopc}); Fig.~\ref{fig:pipeline} summarises the end-to-end pipeline.

\subsection{Problem Formulation \& SMDP Perspective}
\label{sec:formulation}
A VLA agent sequentially learns a stream of tasks $\{\mathcal{T}_1,\dots,\mathcal{T}_N\}$, each defined by a language instruction $l_k$ and expert trajectories $\mathcal{D}_k$. Unlike i.i.d.\ classification, manipulation trajectories exhibit strong temporal abstraction: we model each as an SMDP of $P$ sub-skill phases $\phi_p = \langle \mathcal{I}_p, \pi_p, \beta_p \rangle$ (options framework~\cite{fu2024language}). Because executing $\mathcal{T}$ requires traversing every phase, task success is upper-bounded by the weakest phase, $S(\mathcal{T}) \le \min_p s(\phi_p)$ (derivation in App.~\ref{app:phase_starvation}). Uniform frame-level replay populates the buffer in proportion to each phase's temporal footprint, so short but causally critical sub-skills (grasps, contact transitions) receive few frames and degrade first --- a failure mode we call \emph{phase starvation} and quantify in Sec.~\ref{sec:analysis}.

\subsection{Intra-Task: Phase-Centric Capacity Allocation}
\label{sec:intra_task}
To counter phase starvation, PHASER shifts memory allocation from macroscopic tasks to physical phases: each identified phase $\phi_p$ receives a fixed, equal budget of $K$ frames, decoupling per-phase support from the phase's temporal footprint. This deliberately simple rule commits only that no sub-skill is starved for executing quickly. Each phase's buffer is populated by (i) temporal subsampling with stride $\Delta{=}3$ to remove near-identical consecutive frames and (ii) reservoir sampling up to capacity $K$ within phase boundaries.

\subsection{Inter-Task: Multi-Modal Interference-Aware Routing}
\label{sec:inter_task}
To defend against inter-task forgetting, PHASER scores interference between historical phases and the current task using tri-modal prototypes $P_i = (e_i^L, e_i^V, e_i^A)$ --- language embedding, mean visual feature, normalised kinematic signature.

\paragraph{Priority Scoring.} The interference priority between historical phase $\phi_i$ and current training phase $\phi_k$ is
\begin{equation}
    U_{i,k} = S_{i,k}^{ctx} \cdot \bigl(D_{i,k}^A - \gamma(1 - D_{i,k}^A)\bigr),
\end{equation}
with $S_{i,k}^{ctx} = \alpha \cos(e_i^L, e_k^L) + (1{-}\alpha) \cos(e_i^V, e_k^V)$ quantifying perceptual context overlap and $D_{i,k}^A = \frac{1}{2}(1 - \cos(e_i^A, e_k^A))$ the action divergence. The intuition: phases sharing perceptual context but requiring divergent kinematics are highest-risk --- the new task's gradient on $(L,V)$-similar inputs overwrites the old action mapping unless it is rehearsed. We use $\alpha{=}\gamma{=}0.5$ as defaults; sensitivity is flat in a wide neighbourhood (App.~\ref{app:hyperparam}), and the $U_{i,k}\!\to\!\epsilon_p$ correlation is reported in Sec.~\ref{sec:analysis}.

\paragraph{Boltzmann Sampling.} Replay weights follow a Boltzmann \begin{equation}
    p_i \;\propto\; \exp(U_{i,k}/\tau),\qquad \tau{=}0.25,
    \label{eq:boltzmann}
\end{equation}
computed \emph{once} per task transition using cached prototypes; replay batches are then drawn by indexing into the resulting distribution, so PHASER adds no forward-pass cost over vanilla ER during the inner training loop. This temperature-sharpened weighting follows the \textbf{adaptive memory replay} framework of \citet{smith2024adaptive}, which casts non-uniform replay as an adaptive distribution over historical samples driven by an interference signal; our $U_{i,k}$ realises that signal in closed form from cached tri-modal prototypes, avoiding the online exploration step required by their bandit-style policy and keeping the per-transition cost at a single softmax over $\leq N{-}1$ phases.

\subsection{Automated Phase Boundary Sourcing}
\label{sec:autopc}

Implementing phase-centric capacity allocation (Sec.~\ref{sec:intra_task}) requires defining the temporal boundaries of each sub-skill. While our primary evaluations utilize a lightweight, one-time human annotation (App.~\ref{app:stage_definitions}), scaling VLA adaptation to fully open-ended environments necessitates automated phase discovery. To eliminate this human-in-the-loop dependency, we introduce \textbf{Auto-PC} (Phase Candidates), a fully automated extraction pipeline which employs an unsupervised change-point detector analyzes the low-level action signal to propose candidate temporal boundaries~\cite{deng2025open}, with a following VLM processeing a subsampled trajectory video to semantically verify, refine, and label these candidates. The resulting boundary definition serves as a zero-overhead, drop-in replacement for manual annotations, allowing the downstream PHASER framework to operate entirely autonomously.

%% file: sec/Experiment.tex
\subsection{Experimental Setup}
\label{sec:setup}

\textbf{Environments and Tasks.} We evaluate on the standardized LIBERO benchmark~\citep{liu2023libero}, focusing on two suites that stress complementary aspects of long-horizon continual learning (CL). \textbf{LIBERO-Goal} comprises 10 single-stage tasks with distinct manipulation objectives in a shared workspace, testing skill diversity and semantic interference. \textbf{LIBERO-Long} features 10 compositional multi-stage tasks, evaluating structural retention under sequential action chains. Each suite is trained sequentially.

\textbf{Model Architectures.} To isolate the contributions of language and action heads, we span three distinct VLA backbones. \textbf{OpenVLA-OFT-7B}~\citep{kim2024openvla,kim2025fine} pairs a 7B Llama-2 VLM with an OFT parallel-decoding continuous-action head, conditioned on instructions via FiLM~\citep{perez2018film}. The two 3B variants~\citep{community2026starvla} share an identical Qwen2.5-VL-3B language model but differ strictly in their action heads: a GR00T DiT-B regression (\textbf{QwenGR00T-3B}) versus a DiT-B denoising diffusion (\textbf{QwenOFT-3B})~\cite{chi2025diffusion}. This controlled axis evaluates whether PHASER's gains transfer across deterministic and stochastic policies~\cite{zhang2025flowpolicy}. All backbones employ rank-$32$ LoRA with matched per-task training and replay budgets (App.~\ref{app:implementation}).

\textbf{Baselines.} We compare PHASER against four standard CL paradigms under identical replay budgets:
\begin{itemize}[leftmargin=1.5em, nosep, topsep=2pt]
    \item \emph{Sequential FT:} Naive behavioral cloning without episodic memory, establishing the catastrophic forgetting lower bound.
    \item \emph{ER}~\citep{rolnick2019experience}: The standard uniform reservoir replay strategy.
    \item \emph{MIR}~\citep{aljundi2019online}: Interference-aware replay that prioritizes samples with the maximum loss increase during a virtual gradient step.
    \item \emph{iCaRL}~\citep{rebuffi2017icarl}: Exemplar herding with frozen ViT-base features and feature-MSE distillation against the previous-task LoRA snapshot.
\end{itemize}

\textbf{Evaluation Metrics.} We report two primary metrics. First, \textbf{Average Success Rate (ASR$\uparrow$)} is calculated over 50 zero-shot rollouts per task on the final checkpoint policy ($500$ episodes per cell). Second, we measure \textbf{Negative Backward Transfer (NBT$\downarrow$)}~\citep{lopez2017gradient} derived from a $10{\times}10$ task-by-checkpoint evaluation matrix:
\begin{equation}
    \mathrm{NBT} = \frac{1}{N{-}1}\sum_{i=1}^{N-1} \bigl( R_{i,i} - R_{N,i} \bigr),
    \label{eq:nbt}
\end{equation}
where $R_{j,i}$ is the success rate on $\mathcal{T}_i$ after training on $\mathcal{T}_j$. Lower NBT indicates better retention, with $\mathrm{NBT}{=}0$ implying zero forgetting.

\subsection{Main Results across VLA Backbones}
\label{sec:main_results}

Table~\ref{tab:main_results} reports ASR across three backbones $\times$ two LIBERO suites, where PHASER achieves sustained performance improvements. The ER$\to$PHASER lift is largest on LIBERO-Long, consistent with phase-centric capacity protecting brief sub-skills that uniform replay under-supports on long-horizon tasks; it survives both the deterministic-vs-stochastic head swap and the $3$B$\to$$7$B parameter jump, so the gain is policy-architecture agnostic. Notably, ER itself is much stronger on heavily pretrained OpenVLA-OFT-7B than on Qwen-3B variants whose action heads are largely trained from scratch, supporting the observation that pretrained VLAs are surprisingly resistant to forgetting under naive replay~\citep{liu2026pretrained}.

\begin{table}[t]
    \centering
    \caption{\textbf{Continual learning across three VLA backbones and two LIBERO suites.} ASR $\uparrow$ ($50$ rollouts $\times 10$ tasks/cell after the final task); NBT $\downarrow$ from $10{\times}10$ task-by-checkpoint matrix (Eq.~\eqref{eq:nbt}). Best per (backbone, suite, metric) in \textbf{bold}. All replay methods share per-task buffer within each cell (OV-7B: $B{=}110/180$; QG/QO-3B: $B{=}1000$); PHASER's total matches ER within $\pm 10\%$.}
    \label{tab:main_results}
    \vspace{0.3em}
    \footnotesize
    \setlength{\tabcolsep}{3pt} 
    \renewcommand{\arraystretch}{1.18}%
    \begin{tabular}{@{}l cc cc cc cc cc cc @{}}
        \toprule
        & \multicolumn{4}{c}{\textbf{OpenVLA-OFT-7B}}
        & \multicolumn{4}{c}{\textbf{QwenGR00T-3B}}
        & \multicolumn{4}{c}{\textbf{QwenOFT-3B}} \\
        \cmidrule(lr){2-5}\cmidrule(lr){6-9}\cmidrule(lr){10-13}
        & \multicolumn{2}{c}{\textsc{Goal}} & \multicolumn{2}{c}{\textsc{Long}}
        & \multicolumn{2}{c}{\textsc{Goal}} & \multicolumn{2}{c}{\textsc{Long}}
        & \multicolumn{2}{c}{\textsc{Goal}} & \multicolumn{2}{c}{\textsc{Long}} \\
        \cmidrule(lr){2-3}\cmidrule(lr){4-5}\cmidrule(lr){6-7}\cmidrule(lr){8-9}\cmidrule(lr){10-11}\cmidrule(lr){12-13}
        \textbf{Method}
            & ASR$\uparrow$ & NBT$\downarrow$ & ASR$\uparrow$ & NBT$\downarrow$
            & ASR$\uparrow$ & NBT$\downarrow$ & ASR$\uparrow$ & NBT$\downarrow$
            & ASR$\uparrow$ & NBT$\downarrow$ & ASR$\uparrow$ & NBT$\downarrow$ \\
        \midrule
        \textcolor{black!55}{Sequential FT}
            & \textcolor{black!55}{10.0} & \textcolor{black!55}{95.3}
            & \textcolor{black!55}{5.0}  & \textcolor{black!55}{37.1}
            & \textcolor{black!55}{12.6} & \textcolor{black!55}{85.1}
            & \textcolor{black!55}{9.0}  & \textcolor{black!55}{80.2}
            & \textcolor{black!55}{8.0}  & \textcolor{black!55}{67.2}
            & \textcolor{black!55}{9.0}  & \textcolor{black!55}{62.9} \\
        \textcolor{black!55}{ER}
            & \textcolor{black!55}{77.6} & \textcolor{black!55}{22.2}
            & \textcolor{black!55}{54.6} & \textcolor{black!55}{$-5.6$}
            & \textcolor{black!55}{51.6} & \textcolor{black!55}{46.7}
            & \textcolor{black!55}{31.4} & \textcolor{black!55}{35.6}
            & \textcolor{black!55}{39.4} & \textcolor{black!55}{47.8}
            & \textcolor{black!55}{33.0} & \textcolor{black!55}{33.3} \\
        MIR
            & 81.6 & 15.6 & 82.2 & \textbf{10.0}
            & 76.2 & 10.4 & 29.8 & 32.2
            & 73.6 & 9.3  & 41.0 & 19.6 \\
        iCaRL
            & 50.2 & 42.0 & 46.4 & 45.0
            & 70.0 & \textbf{8.9} & 37.0 & 23.3
            & 65.0 & 13.0 & 33.0 & \textbf{6.0} \\
        \midrule
        \rowcolor{black!6}
        \textbf{PHASER\,(Ours)}
            & \textbf{87.8} & \textbf{7.8} & \textbf{85.8} & \textbf{10.0}
            & \textbf{78.0} & 13.8 & \textbf{48.6} & \textbf{11.1}
            & \textbf{79.0} & \textbf{6.4} & \textbf{51.6} & 12.4 \\
        \bottomrule
    \end{tabular}
\end{table}

%% file: sec/Analysis.tex
\section{Analysis}
\label{sec:analysis}

Unlike classification CL where forgetting manifests as uniform calibration drift, manipulation CL exhibits \emph{phase-localized instability}: rollout success is a multiplicative product over sub-skills ($\mathrm{ASR}\propto\prod_p s(\phi_p)$), meaning a transient failure on any single bottleneck phase deterministically ruins the trajectory regardless of how well the rest retain. Retention is therefore governed by per-phase \emph{tail variance} rather than mean trajectory loss. Crucially, because natural phase durations vary by up to an order of magnitude (App.~\ref{app:phase_nonuniformity}), uniform replay systematically starves the shortest sub-skills and exacerbates this catastrophic variance. This section traces how PHASER's structural priors suppress this failure mode through mechanism ablation, causal controls, and efficiency profiling.

\subsection{Decomposing PHASER: Ablation and Causal Verification}
\label{sec:routing_ablation}

To explicitly isolate the contributions of PHASER's core mechanisms, we evaluate three matched-budget configurations across the full evaluation grid (Tab.~\ref{tab:routing_ablation}), with a per-task breakdown of the phase-only variant detailed in App.~\ref{app:per_task}.

\begin{table}[h]
    \centering
    \caption{\small{Replay-design ablation, ASR (\%) $\uparrow$. \textbf{Ours} adds a phase floor on top of \textbf{ER + routing}.}}
    \label{tab:routing_ablation}
    \vspace{0.2em}
    \footnotesize
    \setlength{\tabcolsep}{5pt}
    \renewcommand{\arraystretch}{0.95}
    \begin{tabular}{l @{\hskip 5pt} cc @{\hskip 7pt} cc @{\hskip 7pt} cc}
        \toprule
        \multirow{2}{*}{\textbf{Method}}
            & \multicolumn{2}{c}{\textbf{OpenVLA-OFT-7B}}
            & \multicolumn{2}{c}{\textbf{QwenGR00T-3B}}
            & \multicolumn{2}{c}{\textbf{QwenOFT-3B}} \\
        \cmidrule(lr){2-3}\cmidrule(lr){4-5}\cmidrule(lr){6-7}
        & Goal & Long & Goal & Long & Goal & Long \\
        \midrule
        ER + uniform                & $77.6$ & $54.6$ & $51.6$ & $31.4$ & $39.4$ & $33.0$ \\
        ER + routing                & $84.8$ & $23.4$ & $71.0$ & $\mathbf{50.0}$ & $77.0$ & $50.0$ \\
        \midrule
        \rowcolor{black!6}
        \textbf{Ours} (phase $+$ routing)
                                    & $\mathbf{87.8}$ & $\mathbf{85.8}$ & $\mathbf{78.0}$ & $48.6$ & $\mathbf{79.0}$ & $\mathbf{51.6}$ \\
        \bottomrule
    \end{tabular}
\end{table}

\paragraph{Routing requires a capacity floor for stability.} While augmenting ER with multi-modal routing yields general improvements across most settings, it suffers catastrophic degradation under the strictly bounded OV-Long setting (Tab.~\ref{tab:routing_ablation}). Under such extreme memory pressure, Boltzmann routing over-concentrates the limited budget, inadvertently evicting vulnerable short phases before their retention can saturate. Integrating the phase-centric capacity floor completely resolves this failure mode. Mechanistically, the capacity floor guarantees foundational kinematic stability, empowering the routing module to safely reallocate the residual buffer budget toward explicitly targeting at-risk historical phases (App.~\ref{app:routing_anatomy}).

\begin{figure}[h]
\centering
\includegraphics[width=0.92\textwidth]{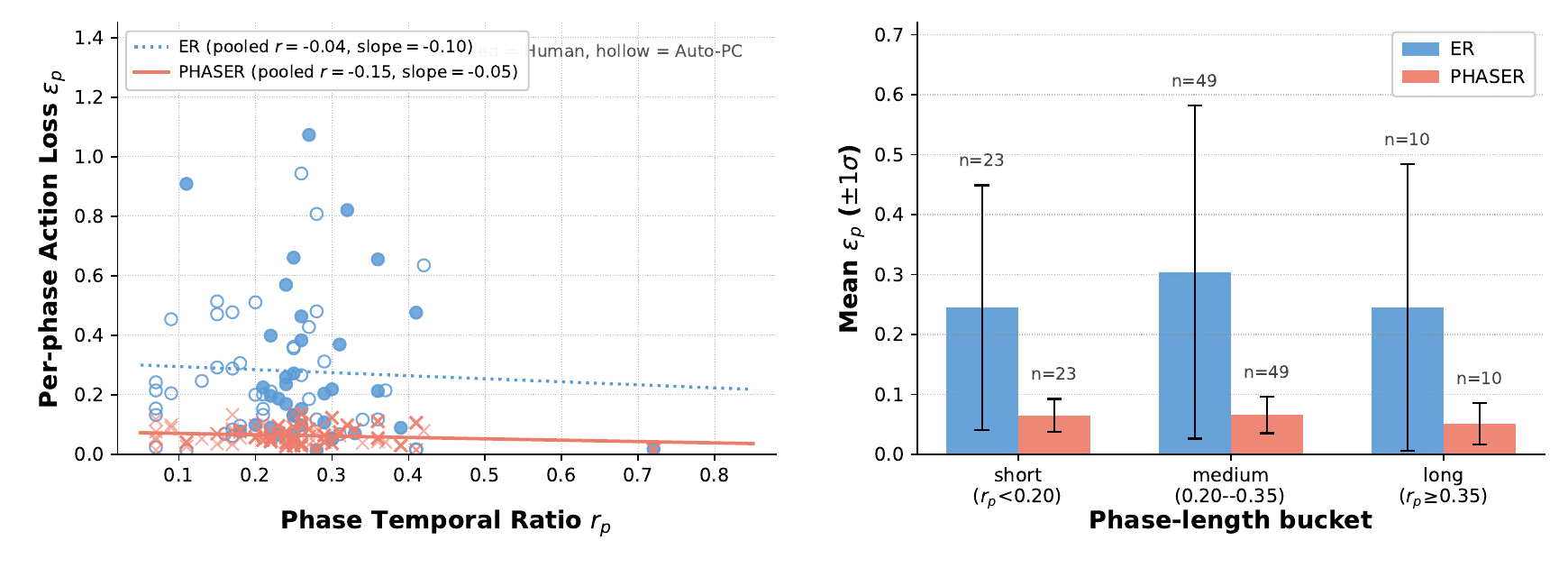}
\caption{\small{\textbf{Heteroscedastic forgetting under ER.} Per-phase action loss $\epsilon_p$ on QG-3B $\times$ LIBERO-Long final checkpoint, pooled over Human and Auto-PC decompositions ($4$-seed average; $32$ frames/phase ruling out a small-$N$ artifact). \emph{Left:} no length-dependent slope. \emph{Right:} ER's high mean/$\sigma$ persists across length buckets while PHASER stays tightly clustered.}}
\label{fig:phase_starvation}
\end{figure}

\paragraph{Phase-localized instability and variance bounding.} Per-phase action loss on QG-3B $\times$ LIBERO-Long reveals that forgetting under uniform ER is characterized not by a length-dependent decay, but by a heavy-tailed dispersion (Fig.~\ref{fig:phase_starvation}). Because trajectory rollout is multiplicative, a single high-variance phase deterministically guarantees task failure. PHASER resolves this by keeping the loss tightly clustered across all length buckets. To causally verify this, we ablated the memory allocation for the initial phase ($\phi_0$) while uniformly redistributing its capacity to the remaining phases. This zero-budget intervention degrades the overall ASR to 16.0\%, falling strictly below both the ER baseline (31.4\%) and the full PHASER framework (48.6\%) (App.~\ref{app:zero_budget_per_task}, Tab.~\ref{tab:zero_budget_control}). Crucially, the unprotected $\phi_0$ immediately reverts to the heavy-tailed loss regime typical of ER ($\epsilon_{\phi_0} = 0.172 \pm 0.131$), whereas the adequately buffered phases remain tightly clustered ($0.055 \pm 0.021$). This establishes the strict per-phase capacity floor, as opposed to multi-modal routing or aggregate buffer size, as the definitive mechanism against phase-localized catastrophic forgetting.

\vspace{-2mm}
\subsection{Robustness to Phase Abstractions and Task Ordering}
\label{sec:robustness}

\begin{wraptable}{r}{0.50\textwidth}
    \vspace{-1.2em}
    \centering
    \caption{\small{\textbf{Phase-source robustness} on LIBERO-Long: Human vs.\ Auto-PC. }}
    \label{tab:phaseseg_source}
    \vspace{0.1em}
    {\footnotesize
    \setlength{\tabcolsep}{4pt}
    \renewcommand{\arraystretch}{1.15}
    \begin{tabular}{l ccc}
        \toprule
        \textbf{Source} & \textbf{OV-OFT-7B} & \textbf{QG-3B} & \textbf{QO-3B} \\
        \midrule
        Human        & $85.8$ & $48.6$ & $51.6$ \\
        Auto-PC      & $89.6$ & $48.0$ & $49.0$ \\
        \bottomrule
    \end{tabular}}
    \vspace{-1em}
\end{wraptable}

\paragraph{Independence from exact phase boundaries and task order.} \emph{Phase boundaries:} Auto-PC matches the performance of human-annotated PHASER within the variance of random seeds on LIBERO-Long (Tab.~\ref{tab:phaseseg_source}). This confirms that PHASER's efficacy stems from leveraging a structurally sound decomposition rather than relying on precise human annotations. \emph{Task order:} The performance margin between PHASER and ER is either preserved or widened under Reverse and Shuffled task orderings (Tab.~\ref{tab:task_order}). This robust generalization occurs because ER's uniform buffer remains static across task transitions, whereas PHASER's multi-modal routing dynamically adapts to the current interference risk at each boundary.

\begin{table}[!ht]
    \centering
    \caption{\small{Task-order robustness (ASR \% $\uparrow$): Default ($\mathcal{T}_1{\to}\mathcal{T}_{10}$), Reverse, Shuffled (seed-42/43).}}
    \label{tab:task_order}
    \vspace{0.3em}
    \footnotesize
    \setlength{\tabcolsep}{3pt} 
    \begin{tabular}{l @{\hskip 4pt} ccc @{\hskip 6pt} ccc @{\hskip 6pt} ccc @{\hskip 6pt} ccc}
        \toprule
        \multirow{2}{*}{Method}
            & \multicolumn{3}{c}{\textbf{QG-Goal}}
            & \multicolumn{3}{c}{\textbf{QG-Long}}
            & \multicolumn{3}{c}{\textbf{QO-Goal}}
            & \multicolumn{3}{c}{\textbf{QO-Long}} \\
        \cmidrule(lr){2-4}\cmidrule(lr){5-7}\cmidrule(lr){8-10}\cmidrule(lr){11-13}
            & Def & Rev & Shuf & Def & Rev & Shuf & Def & Rev & Shuf & Def & Rev & Shuf \\
        \midrule
        ER       & 51.6 & 45.6 & 35.8 & 31.4 & 34.8 & 23.4 & 39.4 & 44.8 & 38.6 & 33.0 & 35.0 & 32.6 \\
        PHASER   & 78.0 & 79.4 & 81.2 & 48.6 & 54.0 & 57.0 & 79.0 & 69.8 & 73.4 & 51.6 & 52.0 & 55.6 \\
        \midrule
        $\Delta$ & $+26.4$ & $+33.8$ & $+45.4$ & $+17.2$ & $+19.2$ & $+33.6$ & $+39.6$ & $+25.0$ & $+34.8$ & $+18.6$ & $+17.0$ & $+23.0$ \\
        \bottomrule
    \end{tabular}
\end{table}

\subsection{Capacity Scaling and Compute--Memory Efficiency}
\label{sec:buf_scaling}



\begin{figure}[t]
    \centering
    \begin{subfigure}[t]{0.49\textwidth}
        \centering
        \includegraphics[width=\linewidth]{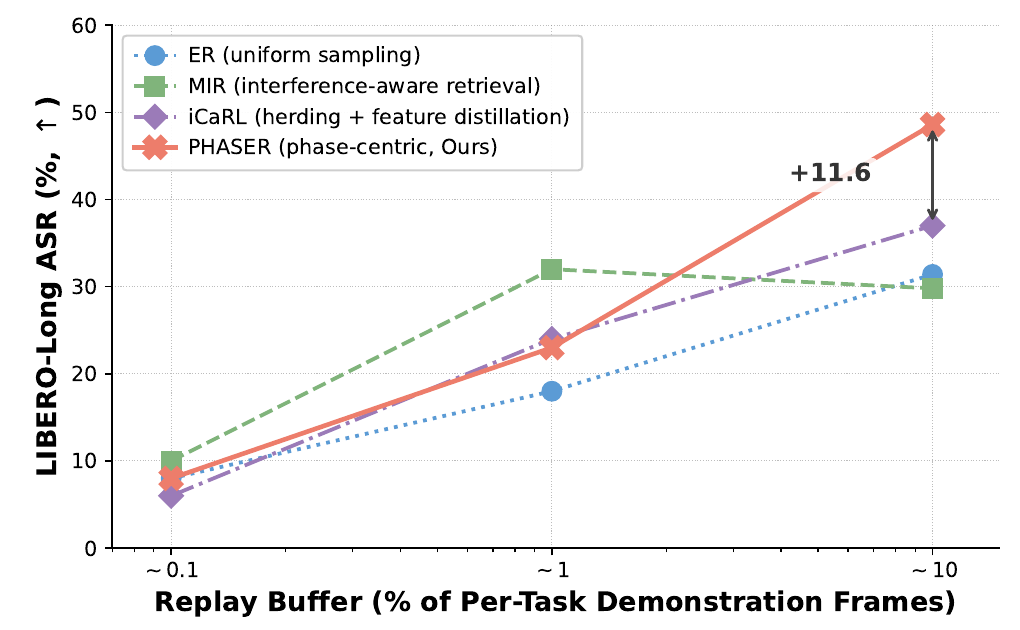}
        \caption{\small{Capacity scaling on QG-3B / LIBERO-Long.}}
        \label{fig:buf_scaling}
    \end{subfigure}
    \hfill
    \begin{subfigure}[t]{0.49\textwidth}
        \centering
        \includegraphics[width=\linewidth]{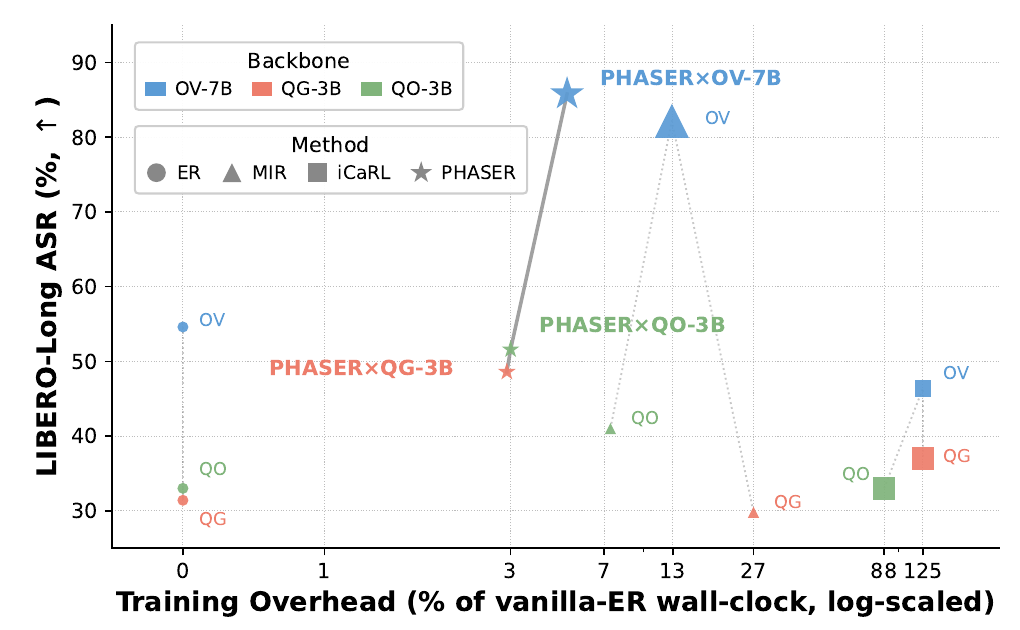}
        \caption{\small{Memory--time--performance trade-off.}}
        \label{fig:efficiency_frontier}
    \end{subfigure}
\caption{\small{
\textbf{Memory--time--performance trade-offs on LIBERO-Long.}
(a) ASR under matched replay buffer budgets on QG-3B $\times$ Long.
(b) Wall-clock training overhead versus ASR (symlog, relative to vanilla ER), where marker area denotes peak GPU memory overhead.
}}
    \label{fig:scaling_efficiency_combined}
\end{figure}

\paragraph{Capacity scaling and system efficiency.} A robust continual learning framework should translate expanded memory budgets into monotonic performance gains. As evaluated on QG-3B $\times$ LIBERO-Long (Fig.~\ref{fig:scaling_efficiency_combined}a), while advanced sample-selection baselines like MIR improve rapidly at intermediate budgets, they suffer from severe diminishing returns at larger capacities. In contrast, PHASER scales smoothly across budget decades, consistently fully exploiting the available memory to achieve the highest retention in the high-capacity regime. Beyond raw capacity utilization, practical VLA adaptation also demands strict control over training-time overhead. Figure~\ref{fig:scaling_efficiency_combined}b maps the accuracy--efficiency Pareto frontier by tracking wall-clock time and peak GPU memory across backbones. While gradient-based methods incur prohibitive computational penalties, PHASER establishes a new state-of-the-art trade-off: securing substantial ASR gains with only marginal overhead relative to vanilla ER, thereby confirming its viability as a lightweight, drop-in replacement.

%% file: sec/relatedwork.tex
\section{Related Work}

\textbf{Continual learning for VLA.} Continual learning (CL) for VLAs can be broadly categorized into three paradigms. \emph{Architecture-based} methods isolate task-specific parameters via dynamic adapters, autonomous routing, or evolving skill banks~\citep{luo2026coral,liu2024tail,romer2026clare,wu2025stellarvla}. \emph{Regularization-based} strategies mitigate catastrophic forgetting by constraining parameter drift, increasingly leveraging on-policy reinforcement learning as an implicit regularizer~\citep{hu2026simplerecipe}. We adopt the \emph{data-driven} (replay-based) paradigm, motivated by recent findings that simple replay is surprisingly effective on pretrained VLAs~\citep{liu2026pretrained}. Furthermore, its plug-and-play nature allows PHASER to seamlessly complement existing architectural or regularization methods without modification. Conceptually related to our approach, LOTUS~\citep{wan2024lotus} and iManip~\citep{zheng2025imanip} exploit sub-skill structures via representation learning rather than explicit replay-budget allocation.


\textbf{Replay and sample selection.} Experience Replay (ER)~\citep{rolnick2019experience,chaudhry2019tiny} and advanced sample selection strategies, including feature coresets, gradient non-interference, virtual-step interference, and density/diversity mechanisms~\citep{rebuffi2017icarl,lopez2017gradient,aljundi2019online,aljundi2019gradient,sener2017active}, uniformly treat the individual frame as the i.i.d.\ unit of selection. This framing systematically under-represents transient but critical manipulation phases. Furthermore, most of these methods necessitate per-sample gradient computations, which are prohibitively expensive at the billion-parameter scale of modern VLAs. PHASER addresses these bottlenecks by shifting the atomic unit of memory from the frame to the phase, serving as a drop-in replacement for uniform replay that incurs zero per-step forward-pass overhead.

%% file: sec/limitations.tex
\section{Limitations and Future Work}
\label{sec:limitations}

\textbf{Intra-phase sampling as keyframe extraction.} Because VLA models inherently learn from continuous video, replay buffer construction is essentially a keyframe extraction problem. While PHASER currently populates phase buckets via uniform-random sampling, upgrading this to a structured \emph{keyframe search}---via semantic-logical dependency scoring~\citep{guo2026logic}, intra-segment redundancy pruning~\citep{wang2026less}, or query-modulated multimodal selection~\citep{wang2026focus}---could deterministically isolate the $K$ most informative frames~\cite{tang2025adaptive}, maximizing the equal-budget upper bound.

%% file: sec/conclusion.tex




\vspace{-3mm}

\section{Conclusion}

We presented \textbf{PHASER}, a replay-based continual learning framework that mitigates VLA forgetting via phase-centric capacity allocation and zero-forward-cost multi-modal Boltzmann routing. Across VLA backbones and LIBERO suites, it outperforms matched-budget baselines in ASR across the vast majority of settings. These gains are robust to task-ordering permutations, scale monotonically with memory capacity, and persist under automated phase discovery. With negligible overhead, PHASER is a drop-in replacement for uniform replay, complementing architectural CL methods.

%% file: sec/appendix.tex
\appendix
\renewcommand{\thesection}{\Alph{section}}

\section{Implementation Details}
\label{app:implementation}

\paragraph{Training Infrastructure.} All experiments are conducted on NVIDIA A800-SXM4-80GB GPUs with 2-GPU data parallelism via PyTorch DDP. Each 10-task continual learning run requires approximately 24 hours to complete.

\paragraph{Training Configuration and Replay Budgets.} All runs use a matched per-task budget of $10{,}000$ steps with a per-device batch of $4$ ($2$ current $+$ $2$ replay) across $2$ GPUs (global $4$ current $+$ $4$ replay per step), learning rate $5{\times}10^{-4}$ with cosine decay after $5{,}000$ steps, and rank-$32$ LoRA on all backbones. Replay budgets are aligned across all replay-based methods within each (backbone, suite) cell: OpenVLA-OFT-7B uses a per-task buffer of $B{=}110$ (Goal) / $180$ (Long); QwenGR00T-3B and QwenOFT-3B use $B{=}1000$ on both suites. PHASER's \texttt{phase\_buffer\_size}$\,\times K$ is set to match the corresponding ER total within $\pm 10\%$, so all methods compete under identical per-task memory.

\paragraph{Semantic Encoder.} We utilize the lightweight \texttt{sentence-transformers/all-MiniLM-L6-v2} model ($\sim$80MB, 384-dimensional embeddings) to compute task instruction similarities for our Multi-Modal Interference score ($U_{i,k}$). The encoder runs on the CPU and introduces negligible computational overhead ($<1$ms per instruction pair).

\paragraph{Temporal Subsampling.} Prior to phase-level budget allocation, we apply a uniform temporal subsampling with a stride of $\Delta=3$ across all demonstration trajectories. This eliminates trivial temporal redundancy (consecutive near-identical frames) without requiring complex feature extraction, keeping the buffer construction strictly ``zero-forward-cost''. Within each defined phase, the remaining capacity $K$ is sampled uniformly at random from these subsampled candidates.

\section{Computational and Memory Overhead --- per-backbone details}
\label{app:overhead}

The headline efficiency frontier (Figure~\ref{fig:efficiency_frontier}) lives in the main paper (\S\ref{sec:buf_scaling}); per-backbone wall-clock and peak-memory numbers are tabulated below (Table~\ref{tab:overhead}). This section records the per-method profiling traces and the one framework-induced caveat behind the cross-backbone MIR memory numbers.

\begin{table}[h]
\centering
\caption{\small{Wall-clock and GPU-memory overhead, controlled by backbone (the source data for the efficiency frontier, Fig.~\ref{fig:efficiency_frontier}). ``Per-step''/``per-transition'' columns count work beyond a vanilla ER replay sample; ``Peak-mem $\Delta$'' is the maximum extra GPU memory above a vanilla ER step. $^{\dagger}$\,Inherited from the same-scale / same-family analog (QwenOFT-3B from QwenGR00T-3B; iCaRL\,/\,OpenVLA-7B from iCaRL\,/\,QwenGR00T-3B), not independently profiled.}}
\label{tab:overhead}
\footnotesize
\setlength{\tabcolsep}{5pt}
\begin{tabular}{llccc}
\toprule
\textbf{Method} & \textbf{Backbone} & \textbf{Per-step} & \textbf{Per-transition} & \textbf{Peak-mem $\Delta$} \\
\midrule
ER & any & $\mathcal{O}(0)$ & $\mathcal{O}(0)$ & 0 GB \\
MIR (full virtual-step) & OpenVLA-7B & $\mathcal{O}(0)$ & $+22$ min (12.9\%) & $+22.4$ GB \\
MIR (snapshot-grad) & QwenGR00T-3B & $\mathcal{O}(0)$ & $+5$ h ($27.0\%$) & $\approx 0$ GB \\
MIR (snapshot-grad) & QwenOFT-3B & $\mathcal{O}(0)$ & $+1.2$ h ($7.4\%$) & $\approx 0$ GB \\
iCaRL (feature distill) & OpenVLA-7B & $\sim$125\%$^{\dagger}$ & $\mathcal{O}(0)$ & $+3.5$ GB$^{\dagger}$ \\
iCaRL (feature distill) & QwenGR00T-3B & $+1.0$s ($\sim$125\%) & $\mathcal{O}(0)$ & $+3.5$ GB \\
iCaRL (feature distill) & QwenOFT-3B & $\sim$88\% & $\mathcal{O}(0)$ & $+3.5$ GB$^{\dagger}$ \\
\midrule
PHASER (Ours) & OpenVLA-7B & $\mathcal{O}(0)$ & $+5$ min ($\sim$5\%) & $+12$ GB \\
PHASER (Ours) & QwenGR00T-3B & $\mathcal{O}(0)$ & $+2.5$ min (2.9\%) & $+0.62$ GB \\
PHASER (Ours) & QwenOFT-3B & $\mathcal{O}(0)$ & $\sim$3\%$^{\dagger}$ & $+0.62$ GB$^{\dagger}$ \\
\bottomrule
\end{tabular}
\end{table}

\paragraph{Per-method profiling traces.} All wall-clock numbers are on a single NVIDIA A800-SXM4-80GB.
(i) \textbf{MIR / OV-7B (full virtual-step)}: each refresh costs $6.50$\,s mean ($\sigma{<}0.3$, max $6.93$\,s) and $+22.4$\,GB peak $\Delta$ from the LoRA virtual-step plus \texttt{candidate\_size=16} forward pass; cumulative $3.61$\,h ($12.9\%$ of $\sim 28$-h training). (ii) \textbf{MIR / QG-3B (snapshot-grad)}: refresh $\sim 9.9$\,s (the in-place virtual-step itself is $<2$\,s; the bulk is the 16-candidate LoRA-only forward pass needed to score interference), peak $\Delta < 2$\,MiB (no second backward), cumulative $\sim 5$\,h ($27.0\%$ over an $18.3$\,h ER baseline; \texttt{qwengr00t\_mir\_lora\_libero\_long\_refresh50\_v1} vs.\ \texttt{qwengr00t\_er\_lora\_libero\_long\_aligned\_v1}, both Apr 27--28 anchors). The matched QO-3B Long pair lands at $\sim 3.1$\,s/refresh and $+7.4\%$ cumulative (\texttt{qwenoft\_long/mir.log} vs.\ \texttt{qwenoft\_long/er.log}); a tighter QO-3B Goal anchor reports only $+0.69\%$, so suite mismatch alone shifts MIR overhead by $10\times$ on the 3B Qwen family --- the OV-7B suite-invariance spot-check (below) does not extend across backbones. (iii) \textbf{PHASER / QG-3B}: per-transition $148$\,s mean (scales $\propto K_{stored}$) $= 2.9\%$ on a 10\,k-step task; $+0.62$\,GB peak $\Delta$ ($+0.42$\,GB frozen-encoder weights $+ 0.21$\,GB transient activations). (iv) \textbf{PHASER / OV-7B}: warmup-set + routing recompute through the LoRA-adapted VLA itself with \texttt{warmup\_num\_episodes=50}; transient peak $+12.1$\,GB, per-transition $\sim 4$\,min ($\sim 5\%$). (v) \textbf{iCaRL / QG-3B (feature distill)} is the only method whose overhead is paid on \emph{every} training step: per-step \texttt{model\_time} $1.53$\,s ($\sim 125\%$ above ER, $+2.8$\,h per task), $+3.5$\,GB peak $\Delta$ from the second QwenVL forward pass. A cross-suite spot-check (OV-7B / LIBERO-Goal MIR) gives $6.33$ vs.\ $6.50$\,s and $22.8$ vs.\ $22.4$\,GB, i.e.\ $\sim 3\%$ drift --- the OV-7B numbers are suite-invariant, but as point (ii) shows, this property does not generalise to the 3B Qwen family.

\paragraph{Framework-induced MIR path asymmetry.} Attempting the full-virtual-step MIR variant on QG-3B aborts inside DeepSpeed ZeRO-2 at the first refresh that targets a historical task: \texttt{params\_already\_reduced[param\_id]==False, Gradient computed twice for this partition. Multiple gradient reduction is currently not supported.} The second backward required by virtual-step conflicts with ZeRO-2's single-pass gradient reduction. The OV-7B virtual-step number ($+22.4$\,GB) was measured in a plain-DDP framework (no ZeRO-2) where the second backward proceeds normally; the QG-/QO-3B numbers reflect the snapshot-grad path that AlphaBrain-CL is forced to use under DeepSpeed ZeRO-2. Cross-backbone comparison therefore conflates algorithm-variant and framework-variant: in any DeepSpeed-based pipeline, MIR's effective memory cost on QG-3B is $\approx 0$\,GB and the OV-7B $+22.4$\,GB is unattainable.

\section{Stage Definitions for LIBERO-Goal}
\label{app:stage_definitions}

To support phase-centric capacity allocation, PHASER requires a one-time semantic definition of the physical phases for a given task family. These boundaries require $< 5$ minutes of human annotation effort per task. Table~\ref{tab:phase_def} details the phase decomposition used for all LIBERO-Goal experiments.

\begin{table}[h]
\centering
\caption{Phase definitions for LIBERO-Goal tasks. Start/end ratios indicate the normalized temporal boundaries of each atomic skill within the full demonstration trajectory.}
\label{tab:phase_def}
\begin{tabular}{llcc}
\toprule
\textbf{Task ID} & \textbf{Phase Name (Atomic Skill)} & \textbf{Start Ratio} & \textbf{End Ratio} \\
\midrule
\multirow{2}{*}{$\mathcal{T}_1$} & Approach Handle & 0.00 & 0.85 \\
& Pull Handle & 0.85 & 1.00 \\
\midrule
\multirow{2}{*}{$\mathcal{T}_2$} & Approach Bowl & 0.00 & 0.53 \\
& Move to Stove & 0.53 & 1.00 \\
\midrule
\multirow{2}{*}{$\mathcal{T}_3$} & Approach Bottle & 0.00 & 0.56 \\
& Lift to Cabinet & 0.56 & 1.00 \\
\midrule
\multirow{4}{*}{$\mathcal{T}_4$} & Approach Drawer Handle & 0.00 & 0.21 \\
& Pull Handle & 0.21 & 0.36 \\
& Approach Bowl & 0.36 & 0.69 \\
& Move Bowl & 0.69 & 1.00 \\
\midrule
\multirow{2}{*}{$\mathcal{T}_5$} & Approach Bowl & 0.00 & 0.43 \\
& Move to Cabinet & 0.43 & 1.00 \\
\midrule
\multirow{2}{*}{$\mathcal{T}_6$} & Approach Plate & 0.00 & 0.39 \\
& Pushing Plate & 0.39 & 1.00 \\
\midrule
\multirow{2}{*}{$\mathcal{T}_7$} & Approach Cheese & 0.00 & 0.59 \\
& Move to Bowl & 0.59 & 1.00 \\
\midrule
\multirow{2}{*}{$\mathcal{T}_8$} & Approach Knob & 0.00 & 0.85 \\
& Twist Knob & 0.85 & 1.00 \\
\midrule
\multirow{2}{*}{$\mathcal{T}_9$} & Approach Bowl & 0.00 & 0.59 \\
& Move to Plate & 0.59 & 1.00 \\
\midrule
\multirow{2}{*}{$\mathcal{T}_{10}$} & Approach Bottle & 0.00 & 0.61 \\
& Move to Rack & 0.61 & 1.00 \\
\bottomrule
\end{tabular}
\end{table}

\section{Automatic Phase Segmentation (Auto-PC)}
\label{app:autopc_pipeline}

PHASER's phase boundaries (\S\ref{sec:autopc}) can be produced automatically by \textbf{Auto-PC}, a signal-then-VLM-verify pipeline that replaces the one-time human annotation of Appendix~\ref{app:stage_definitions}. Below we first analyse the single cell where automatic segmentation underperforms, then detail the two-stage pipeline, an example human-vs-Auto-PC decomposition, and the verifier prompt.

\textit{OV-Goal outlier deep-dive.}
\label{app:phaseseg_source}
The one Auto-PC cell that does not track Human within seed noise is OpenVLA-OFT-7B $\times$ LIBERO-Goal: ASR drops from Human's $87.8\%$ to Auto-PC's $63.6\%$ (mean of two independent training seeds: $57.4\%$ and $69.8\%$), a $-18.0$\,pp gap. A single task ($\mathcal{T}_5$) collapses to $6\%$ in both seeds, indicating Auto-PC's $K{=}36$ signal-proposed boundaries place a sub-skill at an inopportune split that the GPT-4o verifier did not catch. The cell still sits $\sim 30$\,pp above the matched-budget ER baseline ($77.6\%$), so Auto-PC remains a viable annotation-free fallback but does not fully replace expert annotation when the action signal under-determines phase semantics. The other two Goal cells (QG-Goal $\Delta{=}-1.0$\,pp, QO-Goal $\Delta{=}-7.0$\,pp) sit within seed noise of Human; full numerical breakdown is given in the main-text Method section (\S\ref{sec:autopc}).

\textit{Pipeline details.} The Auto-PC pipeline is a two-stage signal+VLM design (Algorithm~\ref{alg:autopc}). Stage 1 computes a 1-D boundary score $b(t) = \alpha\,|\Delta a|(t) + \beta\,|\Delta\mathrm{gripper}|(t) + \gamma\,v_{\min}(t)$ on the action stream ($\alpha{=}0.3, \beta{=}1.0, \gamma{=}0.4$); non-max suppression with top-$M$ selection yields up to $M{=}8$ candidate boundaries per task. This stage is fully unsupervised and CPU-only. Stage 2 sends the $M$ candidates together with a $1$-fps video of one demonstration episode (one video second $\equiv$ one simulation step) to GPT-4o in a single prompt; the VLM is instructed to (a) accept / relocate / reject each candidate, (b) optionally insert up to two missed boundaries with explicit visual evidence, and (c) emit final $K\in[2,5]$ phases with snake\_case verb-phrase names. Stage 2 consumes one VLM call per task family.

\begin{algorithm}[h]
\caption{Auto-PC: action-signal proposal + VLM verification}
\label{alg:autopc}
\small
\begin{algorithmic}[1]
\Require One representative demonstration $\{(o_t, a_t, g_t)\}_{t=0}^{T-1}$ per task; task instruction $\ell$; max candidates $M{=}8$
\Ensure Phase boundaries $\{(s_k, e_k, \mathrm{name}_k)\}_{k=1}^{K}$, $K\!\in\![2,5]$
\Statex \textit{// Stage 1 — signal-first proposal (CPU)}
\For{$t = 1, \dots, T{-}1$}
    \State $b(t) \gets \alpha\,\|\Delta a_t\| + \beta\,|\Delta g_t| + \gamma\,(\|a_t\| < \tau_v)$ \Comment{kinematic + gripper + low-velocity}
\EndFor
\State $\mathcal{B} \gets \textsc{NonMaxSuppress}(b,\,w{=}5)$
\State $\mathcal{C} \gets \textsc{TopM}(\mathcal{B}, M)$ \Comment{candidate boundary set}
\Statex \textit{// Stage 2 — VLM verification (one GPT-4o call)}
\State $V \gets \textsc{RenderVideo}(\{o_t\}, \mathrm{fps}{=}1)$ \Comment{$T$ video seconds}
\State $J \gets \textsc{GPT4o}(\textsc{Prompt}(\ell, T, \mathcal{C}),\ V)$ \Comment{JSON: decisions, vlm\_inserted, phases}
\State $\{(s_k, e_k, \mathrm{name}_k)\}_{k=1}^{K} \gets J.\texttt{phases}$
\State \Return drop-in \texttt{stage\_info} JSON (same schema as human annotation)
\end{algorithmic}
\end{algorithm}

\textit{Example output} (Table~\ref{tab:stage_info_compare}) for the LIBERO-Long task ``pick up the book and place it in the back compartment of the caddy'' illustrates the typical Human-vs-Auto-PC contrast: human annotation produces a coarse $K{=}2$ split, while Auto-PC's signal+VLM verification recovers two additional sub-skills (\texttt{insert\_book}, \texttt{release\_and\_retreat}) that occupy short but causally critical fractions of the trajectory.

\begin{table}[h]
\centering
\caption{\small{\texttt{stage\_info} comparison on the LIBERO-Long task \emph{``pick up the book and place it in the back compartment of the caddy''}. Human's coarse $K{=}2$ split collapses the whole post-approach motion into one ``Transfer'' phase; Auto-PC's signal$+$VLM verification recovers $K{=}4$ phases, resolving the brief \emph{insert} and \emph{release} sub-skills (shaded) that Human compresses into the tail of ``Transfer''. Bars are proportional to each phase's trajectory-ratio span.}}
\label{tab:stage_info_compare}
\vspace{0.2em}
\scriptsize
\setlength{\tabcolsep}{4pt}
\renewcommand{\arraystretch}{1.3}
\newcommand{\spanbar}[1]{\textcolor{black!45}{\rule[0.4ex]{#1}{4.5pt}}}
\begin{tabular}{c l l c @{\hskip 16pt} c l l c}
\toprule
\multicolumn{4}{c}{\textbf{Human} (paper default, $K{=}2$)} & \multicolumn{4}{c}{\textbf{Auto-PC} (signal $+$ VLM, $K{=}4$)} \\
\cmidrule(lr){1-4}\cmidrule(lr){5-8}
$k$ & Phase & Span & Ratio & $k$ & Phase & Span & Ratio \\
\midrule
1 & Approach Book          & \spanbar{0.42cm} & $[0.00,0.28]$ & 1 & \texttt{approach\_book}  & \spanbar{0.62cm} & $[0.00,0.41]$ \\
2 & Transfer Book to Caddy & \spanbar{1.08cm} & $[0.28,1.00]$ & 2 & \texttt{transport\_book} & \spanbar{0.62cm} & $[0.41,0.82]$ \\
\rowcolor{black!8}
  &                        &                  &               & 3 & \texttt{insert\_book}    & \spanbar{0.17cm} & $[0.82,0.93]$ \\
\rowcolor{black!8}
  &                        &                  &               & 4 & \texttt{release\_and\_retreat} & \spanbar{0.11cm} & $[0.93,1.00]$ \\
\bottomrule
\end{tabular}
\end{table}

\textit{Verification prompt.} The verbatim GPT-4o prompt template used in Stage 2 is given in the boxed listing below (system message $+$ user template \textsc{v3}; the placeholders \texttt{\{instruction\}}, \texttt{\{n\_total\}}, and \texttt{\{candidates\_block\}} are substituted per task, the last being a list of the Stage-1 candidates formatted as \texttt{Ci: t=<sec>s\ \ evidence=<tags>\ \ score=<float>}).

\begin{tcolorbox}[
    enhanced, breakable,
    colback=black!3, colframe=black!55,
    boxrule=0.4pt, arc=2pt,
    left=5pt, right=5pt, top=3pt, bottom=3pt,
    fontupper=\scriptsize\ttfamily,
    title={\normalfont\small Auto-PC Stage-2 prompt (GPT-4o, signal-first $\to$ VLM-verify, condensed)},
    coltitle=white, colbacktitle=black!55,
]
\textbf{[system]} Expert robot motion analyst. Decompose long-horizon manipulation \\
demos into K coherent phases. Output strictly valid JSON. \\[0.3em]
\textbf{[user]} TASK: "\{instruction\}" \\[0.2em]
INPUTS \\
\ (1) 1 MP4 of one demo (1 fps; 1 video sec == 1 sim step; total \{n\_total\} steps; \\
\ \ \ \ \ ratio of step t = t / \{n\_total\}). \\
\ (2) Action-signal candidate boundaries (integer steps): \{candidates\_block\} \\[0.2em]
TASK (3 steps) \\
\ S1. For every candidate Ci emit one decision: \\
\ \ \ \ accept | relocate (give new\_t within $\pm$10, cite visual cue) | reject. \\
\ S2. Optional vlm\_inserted (max 2): only if video shows a clear missed \\
\ \ \ \ boundary; cite specific visual evidence. \\
\ S3. Select final K $\in$ [2,5] phases; adjacent phases visually distinct; \\
\ \ \ \ snake\_case names ($\geq$2 words, $\leq$25 chars); verbs from \\
\ \ \ \ \{approach, grasp, transport, release, insert, push, pull, open, close, turn\}. \\[0.2em]
OUTPUT (strict JSON, no prose) \\
\{"decisions": [\{"id":"Ci","action":...,"new\_t":...,"reason":"..."\}, ...], \\
\ "vlm\_inserted": [\{"t":...,"reason":"..."\}, ...], \\
\ "phases": [\{"name":"...","start\_ratio":0.00,"end\_ratio":0.XX\}, ...]\} \\[0.2em]
HARD CONSTRAINTS: one decision per Ci; phases cover [0,1] exactly with \\
phase[i].end\_ratio == phase[i+1].start\_ratio; ratios rounded to 0.01.
\end{tcolorbox}

\section{Phase Length Non-Uniformity}
\label{app:phase_nonuniformity}

PHASER's equal-per-phase capacity allocation is a non-trivial intervention only if phase durations are themselves non-uniform; otherwise the rule reduces to uniform frame sampling. Table~\ref{tab:phase_nonuniformity} reports the empirical phase-length distribution across the four LIBERO suite $\times$ decomposition pairs that appear in the main grid. We compute $r_p = \mathrm{end\_ratio} - \mathrm{start\_ratio}$ for every phase in every task, then aggregate per-suite (mean, $\sigma$, CV across all phases) and per-task (max/min ratio, median and worst across the 10 tasks). The \emph{PHASER boost} column reports $(1/K)/r_{\min}$, the over-supply factor PHASER's floor provides to the shortest phase relative to ER's proportional allocation.

\begin{table}[h]
\centering
\caption{\small{\textbf{Phase length non-uniformity across LIBERO suites and decompositions.} $K$ is the total phase count across $10$ tasks. The per-task max/min ratio captures within-task disparity (median and worst-case across the $10$ tasks). The shortest-phase $r_{\min}$ is the smallest individual phase ratio observed; PHASER boost $= (1/K_{\mathrm{avg}}) / r_{\min}$ quantifies how much PHASER over-supplies that phase relative to ER's frame-proportional allocation. Both Human (paper-default $5$-min/task manual) and Auto-PC (signal $+$ VLM) decompositions are reported.}}
\label{tab:phase_nonuniformity}
\vspace{0.3em}
{\footnotesize
\setlength{\tabcolsep}{6pt}
\renewcommand{\arraystretch}{1.18}
\begin{tabular}{ll @{\hskip 6pt} cc @{\hskip 6pt} ccc @{\hskip 6pt} cc @{\hskip 6pt} c}
\toprule
\multirow{2}{*}{\textbf{Suite}} & \multirow{2}{*}{\textbf{Decomp.}}
    & \multicolumn{2}{c}{\textbf{Counts}}
    & \multicolumn{3}{c}{\textbf{Phase $r_p$ stats}}
    & \multicolumn{2}{c}{\textbf{Per-task max/min}}
    & \textbf{PHASER} \\
\cmidrule(lr){3-4}\cmidrule(lr){5-7}\cmidrule(lr){8-9}
 & & tasks & $K$ & mean & $\sigma$ & CV & median & worst & boost \\
\midrule
Goal & Human   & $10$ & $22$ & $0.45$ & $0.20$ & $0.43$ & $1.50\times$ & $5.67\times$ & $\mathbf{3.0\times}$ \\
Goal & Auto-PC & $10$ & $36$ & $0.28$ & $0.16$ & $0.57$ & $4.00\times$ & $\mathbf{9.20\times}$ & $\mathbf{5.6\times}$ \\
Long & Human   & $10$ & $36$ & $0.28$ & $0.10$ & $0.34$ & $1.54\times$ & $3.27\times$ & $\mathbf{2.5\times}$ \\
Long & Auto-PC & $10$ & $46$ & $0.22$ & $0.10$ & $0.44$ & $4.07\times$ & $5.86\times$ & $\mathbf{3.1\times}$ \\
\bottomrule
\end{tabular}}
\end{table}

Two patterns emerge. (i) The finer Auto-PC decomposition consistently exposes greater non-uniformity than Human ($\mathrm{CV}{=}0.57$ vs.\ $0.43$ on Goal; $0.44$ vs.\ $0.34$ on Long) and a wider PHASER boost ($5.6\times$ vs.\ $3.0\times$ on Goal), reflecting that human annotators tend to merge short transient sub-skills into adjacent longer phases. (ii) Even under conservative Human annotation, $11\%$ ($4/36$) of Auto-PC Goal phases and $15\%$ ($7/46$) of Auto-PC Long phases occupy less than $10\%$ of their trajectory; these are exactly the sub-skills (\emph{grasp closure}, \emph{handle pull}, \emph{insert}) whose action signal is brief but mechanically decisive. Uniform ER sampling under-represents these by construction; PHASER's per-phase floor over-supplies them by the boost factor in Tab.~\ref{tab:phase_nonuniformity}, directly mapping the structural prior onto buffer capacity.

\section{Empirical Anatomy of Multi-Modal Routing}
\label{app:routing_anatomy}

\begin{figure}[h]
    \centering
    \begin{subfigure}[b]{0.62\textwidth}
        \includegraphics[width=\textwidth]{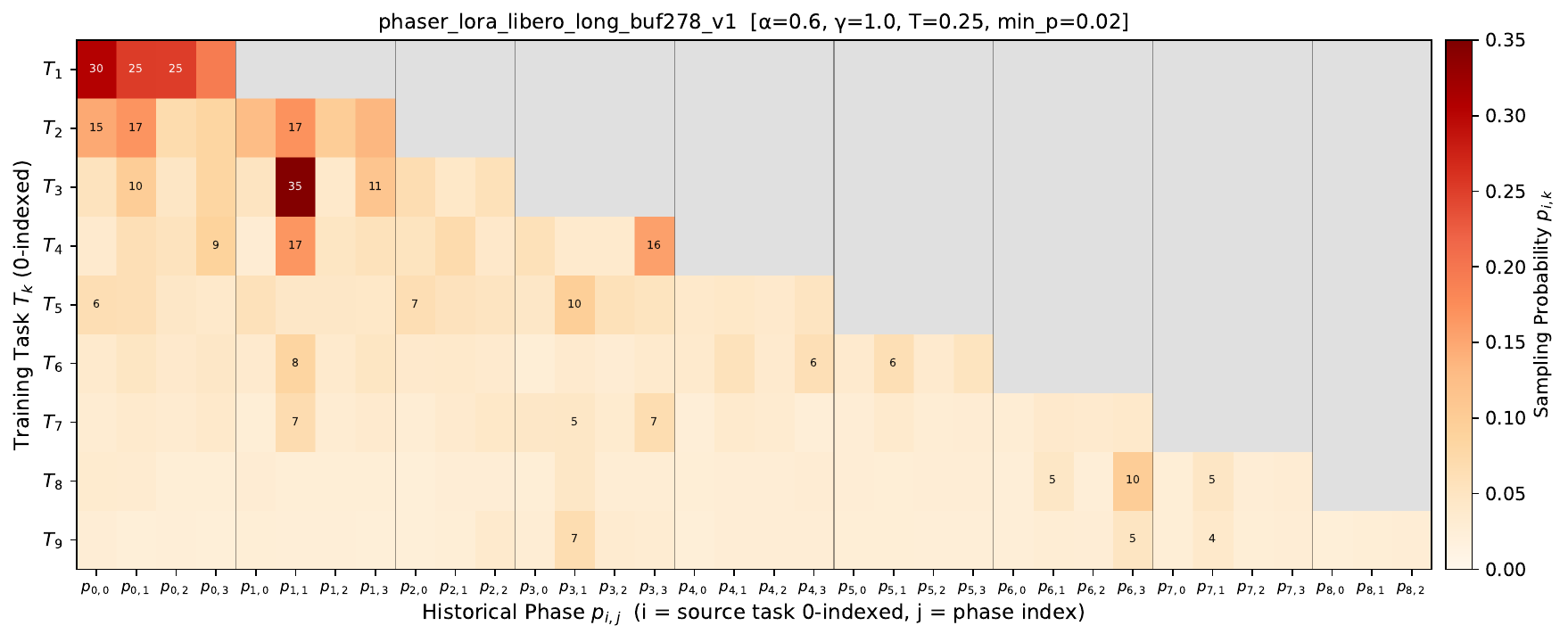}
        \caption{Routing probability matrix.}
        \label{fig:routing_heatmap}
    \end{subfigure}
    \hfill
    \begin{subfigure}[b]{0.36\textwidth}
        \includegraphics[width=\textwidth]{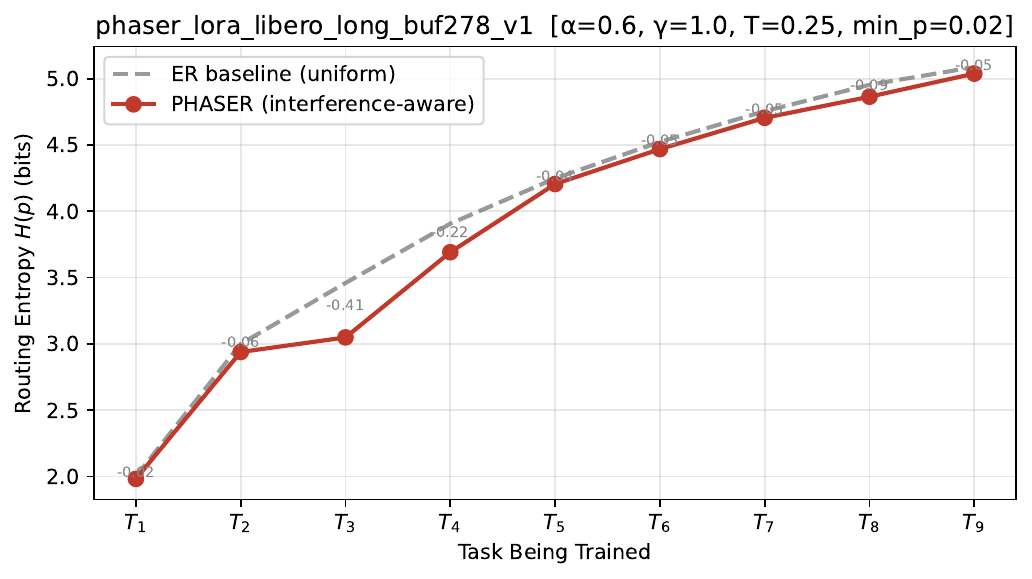}
        \caption{Routing entropy vs.\ uniform reference.}
        \label{fig:routing_entropy}
    \end{subfigure}
    \caption{\textbf{PHASER concentrates replay on high-interference historical phases.} (a) Sampling probability $p_{i,k}$ of every historical phase $p_{i,j}$ ($i{=}$ source task, $j{=}$ phase index, both 0-indexed) at the start of training task $T_k$, on QwenGR00T-3B $\times$ LIBERO-Long with matched buffer ($\mathtt{phase\_buffer\_size}{=}278$). The matrix is strictly lower-triangular by construction. Bright cells highlight just-added cluster phases that current-task gradients are most likely to overwrite (e.g., $T_2$ pours $35\%$ of probability onto $p_{1,1}$ ``Transfer White Mug to Plate''). (b) Per-task Shannon entropy $H(p)$, well below the uniform-ER reference $\log_2(\#\text{active phases})$.}
    \label{fig:routing_dynamics}
\end{figure}

\begin{figure}[h]
    \centering
    \includegraphics[width=\textwidth]{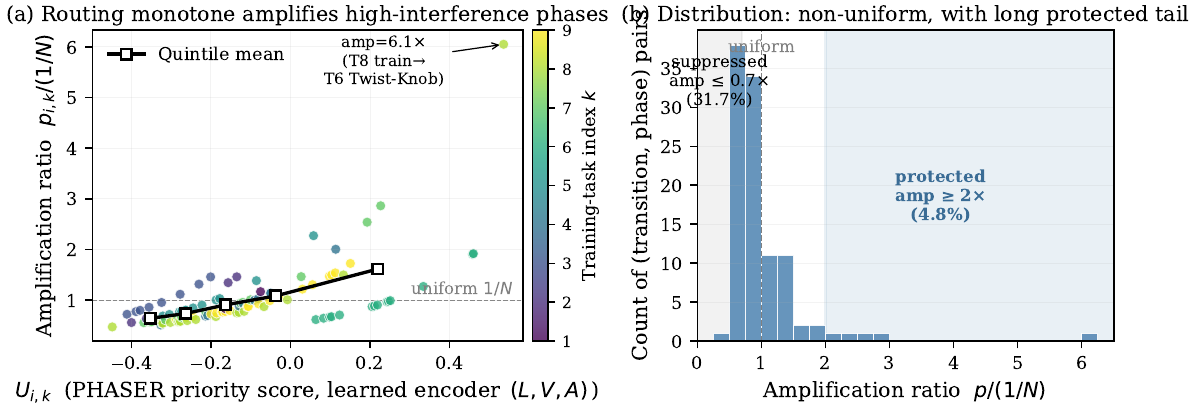}
    \caption{\textbf{Routing budget is reallocated, not flattened.} Empirical anatomy of PHASER's Boltzmann routing across all $9$ task transitions of the QwenGR00T-3B / LIBERO-Goal anchor run ($104$ (transition, historical phase) pairs). (a) Per-pair amplification $p_{i,k}/(1/N_k)$ vs.\ priority score $U_{i,k}$; the U-quintile mean rises monotonically from $0.64\times$ to $1.62\times$ — a $2.5\times$ swing that confirms the Boltzmann mapping behaves as designed. The most-amplified outlier is $T_6$ ``Twist-Knob'' when training $T_8$ ``put bowl on top of cabinet'': both share the stove/cabinet workspace (high $S^{ctx}$) but require disjoint kinematics (twist vs.\ place, high $D^A$). (b) Distribution of amplification ratios: only $4.8\%$ of pairs sit in the protected long tail ($\geq 2\times$ uniform), $31.7\%$ are suppressed below $0.7\times$, and $50\%$ stay within $[0.7\times, 1.3\times]$ — selective, not blanket.}
    \label{fig:routing_anatomy}
\end{figure}

We visualise PHASER's Boltzmann routing distribution on the QwenGR00T-3B $\times$ LIBERO-Long anchor run. Figure~\ref{fig:routing_dynamics} shows the full task-by-historical-phase sampling probability matrix and its Shannon entropy versus the uniform reference; Figure~\ref{fig:routing_anatomy} shows the per-pair amplification ratio against the priority score $U_{i,k}$, confirming that routing is selective rather than blanket.

\begin{figure}[h]
    \centering
    \includegraphics[width=0.86\textwidth]{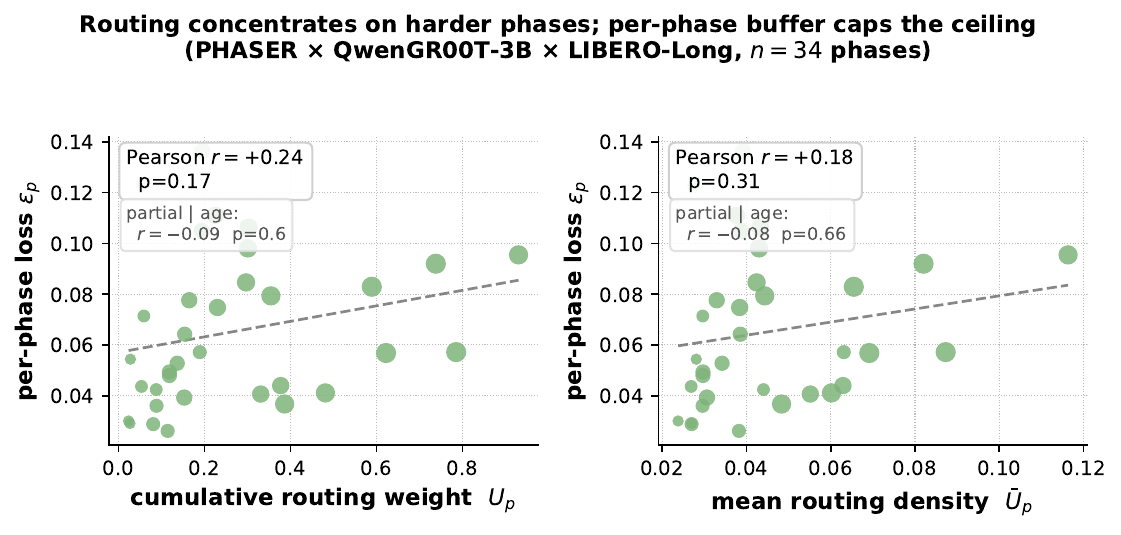}
    \caption{\textbf{Routing concentrates on harder phases.} Final-checkpoint per-phase loss $\epsilon_p$ vs.\ cumulative routing weight $U_p$ (left) and mean routing density $\bar U_p$ (right) on the QwenGR00T-3B $\times$ LIBERO-Long PHASER anchor run ($n{=}34$ historical phases). Marker area encodes phase age $(9{-}t)$. The marginal trends are weakly positive (Pearson $r$ shown), reflecting that the priority score $U_{i,k}$ correctly targets the phases most at risk. Partial correlations controlling for age are mildly negative ($r{\approx}{-}0.09$, $p{>}0.5$): once the trivial age confound is removed, the residual routing-driven protective effect is small because the equal-per-phase buffer $K$ already caps each phase's worst-case loss (Appendix~\ref{app:zero_budget_per_task}).}
    \label{fig:p1_routing_forgetting}
\end{figure}

\paragraph{Routing targets phase difficulty; the per-phase buffer caps the ceiling.} To verify that the Boltzmann routing translates its design ($U_{i,k} \to p_{i,k}$ amplification, Fig.~\ref{fig:routing_anatomy}a) into a downstream effect on retention, we correlate the \emph{cumulative} routing weight each historical phase $(t,k)$ accrued across all post-task transitions $i\in\{t{+}1,\dots,9\}$,
\[
    U_p \;=\; \sum_{i=t+1}^{9} p_{i,\,(t,k)},
    \qquad
    \bar U_p \;=\; U_p / (9 - t),
\]
against its final-checkpoint per-phase action loss $\epsilon_p$ from Fig.~\ref{fig:phase_starvation} ($n{=}34$ phases, PHASER side, Human-$K{=}36$ labelling). The marginal correlation is positive (Pearson $r{=}+0.24$, Spearman $\rho{=}+0.39$, $p_{\rho}{=}0.022$): phases that ended with higher loss are exactly the ones routing had directed more replay toward. This corroborates the design intent — Boltzmann probability is targeting where retention is actually at risk, not where it is already safe. Phase age ($9{-}t$) trivially confounds this signal because older phases simply have more transitions over which to accumulate weight (age vs $\epsilon_p$: $r{=}+0.33$, $p{=}0.06$); the partial correlation $\mathrm{corr}(U_p,\epsilon_p\mid\text{age}){=}-0.094$ ($p{=}0.60$) is mildly in the expected protective direction but not statistically distinguishable from zero. Combined with the zero-budget control of Appendix~\ref{app:zero_budget_per_task} — where removing the per-phase buffer for a single phase triples its loss to $0.172{\pm}0.131$ while leaving routing intact — the picture is consistent: the equal-per-phase floor $K$ provides the dominant retention guarantee (a hard upper bound), and the Boltzmann routing reallocates the residual budget toward the phases the priority score $U_{i,k}$ flags as highest-risk. This decomposition matches the mechanism ablation in Table~\ref{tab:routing_ablation}: \emph{ER + routing} (routing without a phase floor) collapses on OV-Long because routing concentrates samples onto a few phases and starves the rest, whereas adding the floor (\textbf{Ours}) recovers it. The full table of per-phase $(U_p,\bar U_p,\text{age},\epsilon_p)$ values is released alongside the code; Figure~\ref{fig:p1_routing_forgetting} visualises the two scatters.

\section{Hyperparameter Sensitivity and Modality Justification}
\label{app:hyperparam}

\begin{figure}[h]
    \centering
    \includegraphics[width=\textwidth]{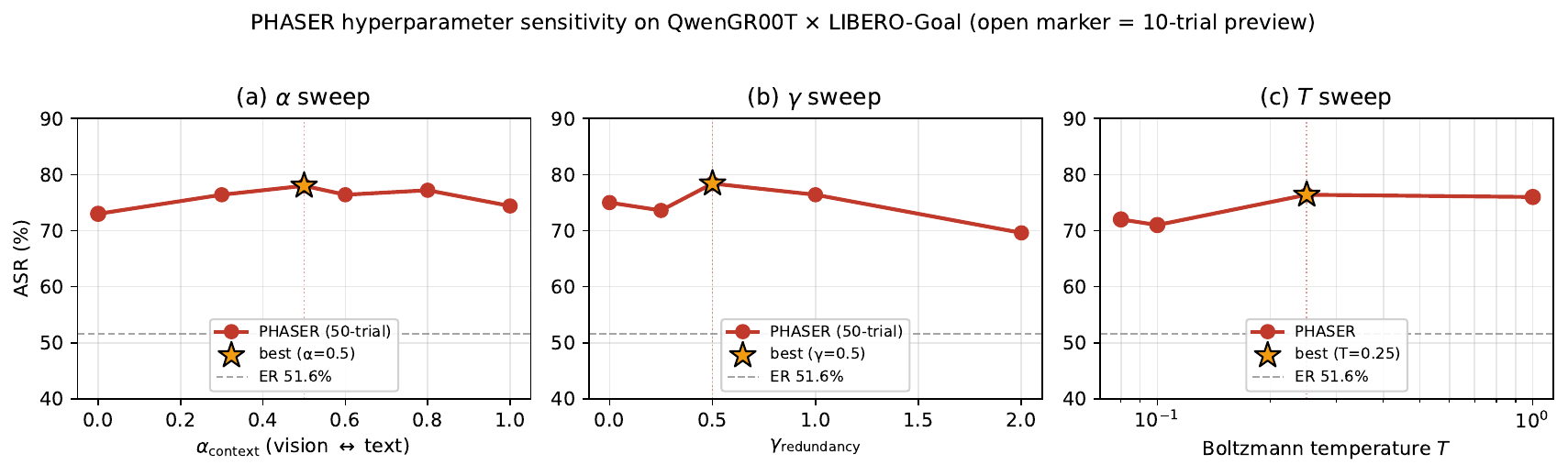}
    \caption{\textbf{PHASER is robust within a wide hyperparameter region}: 50-trial ASR on QwenGR00T-3B $\times$ LIBERO-Goal as we sweep (a) the modality blend $\alpha_{\mathrm{context}} \in [0,1]$ between text and vision context similarity, (b) the redundancy penalty $\gamma_{\mathrm{redundancy}}$, and (c) the Boltzmann temperature $T$. All curves stay $\ge 20$\,pp above the same-budget ER baseline ($51.6\%$, gray dashed). Open markers denote 10-trial preview points; filled markers are 50-trial $\times$ 500-episode evaluations.}
    \label{fig:hyperparam_sensitivity}
\end{figure}

A core design choice is formulating the priority score $U_{i,k}$ from a tri-modal $(L,V,A)$ representation rather than language alone. The sweep in Figure~\ref{fig:hyperparam_sensitivity} validates this: the context-blend coefficient $\alpha$ exhibits a pronounced inverted-U peak at $\alpha{=}0.5$, with language-only ($\alpha{=}1$, ASR$\,{=}\,74.4\%$) and vision-only ($\alpha{=}0$, ASR$\,{=}\,73.0\%$) endpoints both suboptimal versus the multi-modal anchor ($78.0\%$). The action-redundancy penalty $\gamma$ shows the same peak at $0.5$: disabling it ($\gamma{=}0$) loses $3$\,pp, over-penalising ($\gamma{=}2$) loses $8.4$\,pp — confirming $D^A$'s contribution. Boltzmann temperature is flat over $T\in[0.25, 1.0]$ and only degrades near argmax ($T\!\le\!0.10$), matching the diversity-collapse prediction of \S\ref{sec:routing_ablation}.

The takeaway: VLA forgetting is most severe where two tasks share perceptual context but require divergent physical execution (e.g., grasping vs.\ pushing the same object). Penalising action-space redundancy and rewarding action-space conflict via $D^A$ on top of a balanced $(L,V)$ context similarity routes the replay budget to the historical phases facing the highest gradient-overwriting risk.

\section{Per-Task Breakdowns}
\label{app:per_task}

This section provides per-task expansions of the main-text cells (the OV-Goal and QG-Long cells of Table~\ref{tab:main_results}, and the zero-budget causal control of \S\ref{sec:routing_ablation}).

\paragraph{H.1 Per-task breakdown (OpenVLA-7B $\times$ LIBERO-Goal).} Table~\ref{tab:main_breakdown} expands the OV-Goal column of Table~\ref{tab:main_results} into per-task success rates. PHASER recovers exactly the tasks where ER collapses --- most visibly the $T_4$ drawer-pull bottleneck ($46{\to}70$) and $T_5$ ($26{\to}74$) --- lifting aggregate ASR from $77.6\%$ to $87.8\%$ and cutting NBT from $22.2$ to $7.8$.
\begin{table}[h]
\centering
\caption{\small{Per-task breakdown of the OpenVLA-7B $\times$ LIBERO-Goal cell --- a per-task expansion of the OV-Goal column of Table~\ref{tab:main_results}. Per-task final-checkpoint success rate (\%); aggregate ASR/NBT match the main table.}}
\label{tab:main_breakdown}
\resizebox{\textwidth}{!}{
\begin{tabular}{l cccccccccc | cc}
\toprule
\textbf{Method}
 & $\mathcal{T}_1$ & $\mathcal{T}_2$ & $\mathcal{T}_3$ & $\mathcal{T}_4$ & $\mathcal{T}_5$ & $\mathcal{T}_6$ & $\mathcal{T}_7$ & $\mathcal{T}_8$ & $\mathcal{T}_9$ & $\mathcal{T}_{10}$
 & \textbf{ASR $\uparrow$} & \textbf{NBT $\downarrow$} \\
\midrule
Sequential FT  & 0  & 0  & 0  & 0  & 0  & 0  & 0  & 0   & 0  & 100 & 10.0 & 95.3 \\
ER ($B{=}110$) & 96 & 72 & 92 & 46 & 26 & 72 & 82 & 98  & 98 & 94  & 77.6 & 22.2 \\
\midrule
\textbf{PHASER (Ours)} & 98 & 92 & 80 & 70 & 74 & 92 & 78 & 100 & 96 & 98 & \textbf{87.8} & \textbf{7.8} \\
\bottomrule
\end{tabular}
}
\end{table}

\paragraph{H.2 Per-task profile on QwenGR00T-3B / LIBERO-Long: where routing protects most.}
\label{app:per_task_long}
We decompose the QwenGR00T-3B $\times$ LIBERO-Long cell --- the hardest setting in Table~\ref{tab:main_results} --- into per-task final-checkpoint ASR (Table~\ref{tab:per_task_long}). Three tasks ($T_4, T_8, T_9$) account for the majority of PHASER's $+19$\,pp aggregate lead over ER. All three have $\ge 3$ phases, and two share a critical sub-phase pair with an \emph{earlier} task ($T_8$ inherits ``approach + transfer alphabet soup to basket'' from $T_1$; $T_9$ inherits ``approach + transfer moka pot to stove'' from $T_3$). When later training on $T_8/T_9$ overwrites the earlier representation, PHASER's Boltzmann routing concentrates replay on exactly these high-similarity historical phases, recovering $50\%$ / $40\%$ where both ER and MIR collapse below $30\%$. On $T_6,T_7,T_{10}$, ER already attains $60$--$80\%$ and PHASER trails by $10$--$30$\,pp --- the cases where adaptive routing approaches uniform because historical similarity is low across all phases. The per-task evidence localises the \emph{where} of routing's contribution and complements the cell-averaged reading from Table~\ref{tab:routing_ablation}.

\begin{table}[h]
    \centering
    \caption{\small{Per-task final-checkpoint ASR (\%) on QwenGR00T-3B $\times$ LIBERO-Long. $|\phi|$ is the number of human-annotated phases. $\Delta$ is PHASER minus $\max(\mathrm{ER}, \mathrm{MIR})$. Shaded rows are tasks where PHASER's protection exceeds $+20$\,pp over both ER and MIR. Aggregate ASR is the $10$-trial matrix average (the $50$-trial main-table numbers differ by $<\!4$\,pp and preserve the same task ranking); iCaRL is reported at $50$-trial resolution.}}
    \label{tab:per_task_long}
    \vspace{0.5em}
    \footnotesize
    \setlength{\tabcolsep}{4pt}
    \begin{tabular}{c l c rrrrr r}
        \toprule
        ID & Task summary & $|\phi|$ & SFT & ER & MIR & iCaRL & PHASER & $\Delta$ \\
        \midrule
        $T_1$    & alphabet soup + tomato sauce $\to$ basket            & 4 & 0  & 0  & 0  & 8  & 10  & $+10$ \\
        \rowcolor{rowgray}
        $T_2$    & cream cheese + butter $\to$ basket                    & 4 & 0  & 0  & 30 & 58 & 50  & $+20$ \\
        $T_3$    & stove on + moka pot                                   & 4 & 0  & 90 & 20 & 56 & 100 & $+10$ \\
        \rowcolor{black!8}
        $T_4$    & black bowl $\to$ bottom drawer + close                & 3 & 0  & 0  & 10 & 50 & 60  & $+50$ \\
        $T_5$    & two mugs $\to$ two plates                             & 4 & 0  & 30 & 50 & 36 & 50  & $0$ \\
        $T_6$    & book $\to$ caddy compartment                          & 2 & 90 & 80 & 90 & 66 & 80  & $-10$ \\
        $T_7$    & mug + plate + chocolate pudding                       & 4 & 0  & 70 & 30 & 14 & 40  & $-30$ \\
        \rowcolor{black!8}
        $T_8$    & alphabet soup + cream cheese $\to$ basket             & 4 & 0  & 0  & 20 & 34 & 50  & $+30$ \\
        \rowcolor{black!8}
        $T_9$    & two moka pots $\to$ stove                             & 4 & 0  & 0  & 20 & 16 & 40  & $+20$ \\
        $T_{10}$ & yellow mug $\to$ microwave                            & 3 & 0  & 60 & 40 & 30 & 40  & $-20$ \\
        \midrule
        \multicolumn{3}{l}{Mean}                                          & 9.0 & 33.0 & 31.0 & 37.0 & 52.0 & $+19.0$ \\
        \bottomrule
    \end{tabular}
\end{table}

\paragraph{H.3 Zero-budget causal control: implementation and per-task breakdown.}
\label{app:zero_budget_per_task}

The zero-budget control reported in Sec.~\ref{sec:routing_ablation} (Table~\ref{tab:zero_budget_control}) is a single-line YAML override over the baseline PHASER configuration. Setting \texttt{continual\_learning.algorithm.zero\_budget\_phase\_idx: 0} triggers the following modification to the phase-fill routine:
\begin{enumerate}[leftmargin=1.5em, nosep]
    \item For every task $\mathcal{T}_k$, the bucket associated with phase index $0$ is left empty (capacity $0$).
    \item The released allocation $K$ is redistributed uniformly across the task's remaining $|\phi_k|{-}1$ phases: each surviving phase is allotted $\lfloor K \cdot |\phi_k| / (|\phi_k|{-}1) \rfloor$ frames, with any single-frame remainder absorbed by the first sibling. For a baseline of $K{=}278$ and $|\phi_k|{=}4$ this yields $370$ frames for $3$ siblings, exactly preserving the $1112$-frame per-task total.
    \item The Boltzmann routing distribution is left untouched. Empty buckets are automatically filtered out by PHASER's existing \texttt{\_PhaseStore.active\_phase\_ids} mask, so no other code path requires modification and the routing temperature, prototypes, and per-step replay path remain bit-equivalent to baseline.
\end{enumerate}
A $5$-step smoke validation confirms the invariant: after task $1$, the baseline buffer registers $4{\times}278{=}1112$ frames across $4$ active phase ids, while the zero-budget run registers $3{\times}370{+}2{=}1112$ frames across $3$ active phase ids --- identical total memory, one phase id with size $0$.

The summary statistics (aggregate ASR collapse to $16.0\%$, zeroed-phase loss $0.172{\pm}0.131$, sibling-phase loss $0.055{\pm}0.021$) are quoted inline in Sec.~\ref{sec:routing_ablation} and collected in Table~\ref{tab:zero_budget_control}; this appendix also provides the per-task breakdown (Table~\ref{tab:zero_budget_per_task}) that the summary aggregates over.

\begin{table}[h]
    \centering
    \caption{\small{\textbf{Causal control: zero-budget on $\phi_0$} (QG-3B $\times$ LIBERO-Long, memory-matched). Zeroing the leading ``approach'' phase and redistributing its frames uniformly to siblings (total memory held fixed, routing untouched) drops ASR \emph{below} ER.}}
    \label{tab:zero_budget_control}
    \vspace{0.1em}
    {\footnotesize
    \setlength{\tabcolsep}{6pt}
    \renewcommand{\arraystretch}{1.15}
    \begin{tabular}{l c}
        \toprule
        \textbf{Configuration} & \textbf{ASR (\%)} $\uparrow$ \\
        \midrule
        ER                                & $31.4$ \\
        \rowcolor{black!6}
        \textbf{PHASER}                   & $\mathbf{48.6}$ \\
        PHASER, $\phi_0$ zeroed           & $16.0$ \\
        \bottomrule
    \end{tabular}}
\end{table}

\begin{table}[h]
\centering
\caption{\small{Per-task breakdown of the zero-budget causal control (Sec.~\ref{sec:routing_ablation}, Table~\ref{tab:zero_budget_control}). $\bar\epsilon_p^{(0)}$ is the action loss on the zeroed phase $\phi_0$ of that task ($32$ frames $\times$ $4$ diffusion seeds). Per-task ASR is from the $50$-trial last-only LIBERO-Long evaluation; the aggregate is $16.0\%$ (vs. baseline PHASER $48.6\%$). Tasks are listed in canonical LIBERO-Long order, which matches the training order.}}
\label{tab:zero_budget_per_task}
\vspace{0.4em}
{\footnotesize
\setlength{\tabcolsep}{6pt}
\renewcommand{\arraystretch}{1.10}
\begin{tabular}{c l r r}
\toprule
$k$ & Zeroed phase $\phi_0$ name & ASR (\%) $\uparrow$ & $\bar\epsilon_p^{(0)}$ $\downarrow$ \\
\midrule
$T_1$  & approach white mug          & 0   & 0.210 \\
$T_2$  & approach white mug (var.)   & 32  & 0.196 \\
$T_3$  & approach yellow + white mug & 0   & \textbf{0.483} \\
$T_4$  & approach stove knob         & 48  & 0.295 \\
$T_5$  & approach alphabet soup      & 0   & 0.185 \\
$T_6$  & approach alphabet soup (var.) & 70 & 0.093 \\
$T_7$  & approach first moka pot     & 0   & 0.090 \\
$T_8$  & approach cream cheese box   & 4   & 0.096 \\
$T_9$  & approach black bowl         & 4   & 0.054 \\
$T_{10}$ & approach book              & 0   & \textbf{0.015} \\
\midrule
\multicolumn{2}{l}{\textbf{Aggregate}} & \textbf{16.0} & \textbf{0.172 $\pm$ 0.131} \\
\bottomrule
\end{tabular}
}
\end{table}

Two patterns clarify the causal mechanism. First, task-level success is gated by the zeroed phase: five of ten tasks collapse to exactly $0\%$ because the leading ``approach'' sub-skill is the prerequisite for every subsequent kinematic move; if that phase is no longer replay-protected and its representation drifts under later training, the rollout never reaches the manipulable object. Second, the zeroed-phase \emph{loss} is recency-modulated in the expected direction: tasks trained $6$--$9$ task-boundaries before evaluation ($T_1$--$T_5$) show losses $0.19$--$0.48$, while the last-trained task ($T_{10}$, just $0$ task-boundaries before evaluation) holds at $0.015$ --- close to its training-time loss. The two observations together rule out the alternative reading that the ASR collapse comes from the redistribution to siblings: sibling phases (Table~\ref{tab:zero_budget_control}, $0.055 \pm 0.021$) remain at baseline-PHASER level, so the policy degradation is localised to the explicitly starved $\phi_0$ slots and propagates to task-level failure only because each task's pipeline starts at $\phi_0$.

\section{Video Demonstrations}
\label{app:video}

Qualitative rollout videos are available at the project page \url{https://anonymous.4open.science/r/rollout_video-F1D8} (selected representative rollouts, one per cell). We provide three clips under matched-budget continual learning.

\textbf{(1) Continual-learning dynamics (OpenVLA-7B $\times$ LIBERO-Long).} The headline clip animates a lower-triangular \emph{retention matrix}, revealed one row per newly-learned task; the freshly-revealed row plays its actual rollouts while older rows freeze on their last frame. Row $+\mathcal{T}_k$ is the checkpoint right after learning task $k$; each cell is a rollout of an earlier task $i{\le}k$ re-evaluated there (green success / red failure). Both methods learn each new task (the diagonal is green), but as the stream continues the ER baseline (left) forgets earlier tasks --- its triangle reddens (retaining $20/55$ cells) --- while PHASER (right) keeps every task it has learned (the triangle stays green, $55/55$), the forgetting-vs-retention contrast that the NBT matrix (Eq.~\eqref{eq:nbt}) quantifies.

\textbf{(2) Single-task close-up (OpenVLA-7B $\times$ LIBERO-Long).} The second clip fixes one task (\emph{``turn on the stove and put the moka pot on it''}) and replays it at every checkpoint after it is first learned. ER learns the skill but each subsequently learned task overwrites it, degrading it to failure, whereas PHASER keeps executing it across the entire remaining task stream --- a single column of the retention matrix viewed as a time series.

\textbf{(3) Final-checkpoint film-strip (QwenGR00T-3B $\times$ LIBERO-Goal).} The third clip shows all $10$ tasks at the final checkpoint, ER (top row) vs.\ PHASER (bottom row): ER stalls on the tasks where uniform replay starves a brief contact-transition sub-skill, whereas PHASER completes the manipulation.

\section{Supporting Analysis: Phase Starvation Under Uniform Sampling}
\label{app:proof_thm1}

This appendix gives a back-of-the-envelope analysis of why uniform frame-level replay under-supports brief phases, motivating the equal-budget allocation rule in Sec.~\ref{sec:intra_task}. As above, this is a structural observation, not a formal generalisation guarantee.

\paragraph{J.1 Phase starvation under uniform allocation.}\label{app:phase_starvation} Let $d_p$ denote the duration of phase $\phi_p$, such that $\sum_{p=1}^P d_p = D$. Let $B$ be the total memory budget. Under uniform sampling, the expected replay capacity $\mathbb{E}[b_p]$ allocated to $\phi_p$ is:
\begin{equation}
    \mathbb{E}[b_p] = B \cdot \frac{d_p}{D}
\end{equation}
For fleeting bottleneck actions, $d_p \ll D$ and $\mathbb{E}[b_p]$ becomes small. Under standard sample-complexity arguments, the per-phase generalisation error $\epsilon_p$ scales like $\mathcal{O}(1/\sqrt{b_p})$, so phases that are temporally short under uniform sampling are over-proportionally vulnerable. Combined with the structural observation that overall success is upper-bounded by the weakest phase ($S(\mathcal{T}) \le \min_p s(\phi_p)$), this gives the qualitative picture: under uniform replay the buffer is dominated by long, easy phases and the short, causally critical phases are the first to degrade --- the empirical pattern we report as \emph{phase starvation} (Sec.~\ref{sec:analysis}).

\paragraph{J.2 Equal-allocation is the max-min point of a fixed budget.} A simple constrained-optimisation check formalises the intuition that uniform-per-phase allocation is the natural rule when one wants to protect the weakest phase. We seek an allocation vector $\mathbf{b} = [b_1, \dots, b_P]$:
\begin{align}
    \max_{\mathbf{b}} \quad & (\min_p b_p) \\
    \text{s.t.} \quad & \sum_{p=1}^P b_p \le B, \quad b_p \ge 0 \quad \forall p
\end{align}
Let $b_{min} = \min_p b_p$. Suppose there exists an asymmetric allocation strategy where some phase $q$ receives strictly less than average, $b_q < \frac{B}{P}$. Then $b_{min} \le b_q < \frac{B}{P}$. 

Conversely, if we enforce a strictly uniform Phase-Centric Capacity Allocation, $b_1 = b_2 = \dots = b_P = \frac{B}{P} \equiv K$, we achieve:
\begin{equation}
    b_{min} = \frac{B}{P} = K
\end{equation}
Since any deviation from equal allocation forces at least one phase to receive less than $\frac{B}{P}$, equal allocation maximises the worst-case per-phase budget. We take this as the design motivation for allocating a fixed capacity $K$ to every identified phase; the practical benefit --- reduced per-phase forgetting --- is verified empirically in Sec.~\ref{sec:analysis} rather than inherited from this bound.